%% file: main.tex
\ifdefined\pdfminorversion
\fi
\documentclass[cameraready]{mac_automl_xmu_blue}

\makeatletter
\def\input@path{{content/}}
\makeatother
\graphicspath{{content/}}

\usepackage[toc,page,header]{appendix}

\input{content/resources/packages}

\input{content/resources/math_macro}
\DeclareCaptionJustification{fullparagraph}{\leftskip=0pt\rightskip=0pt\parfillskip=0pt plus .2\linewidth}
\setleftheadercontent{%
  \headerlogospace{1.6mm}%
  \raisebox{-0.05\height}{\includegraphics[height=13mm,trim={0bp 0bp 0bp 0bp},clip]{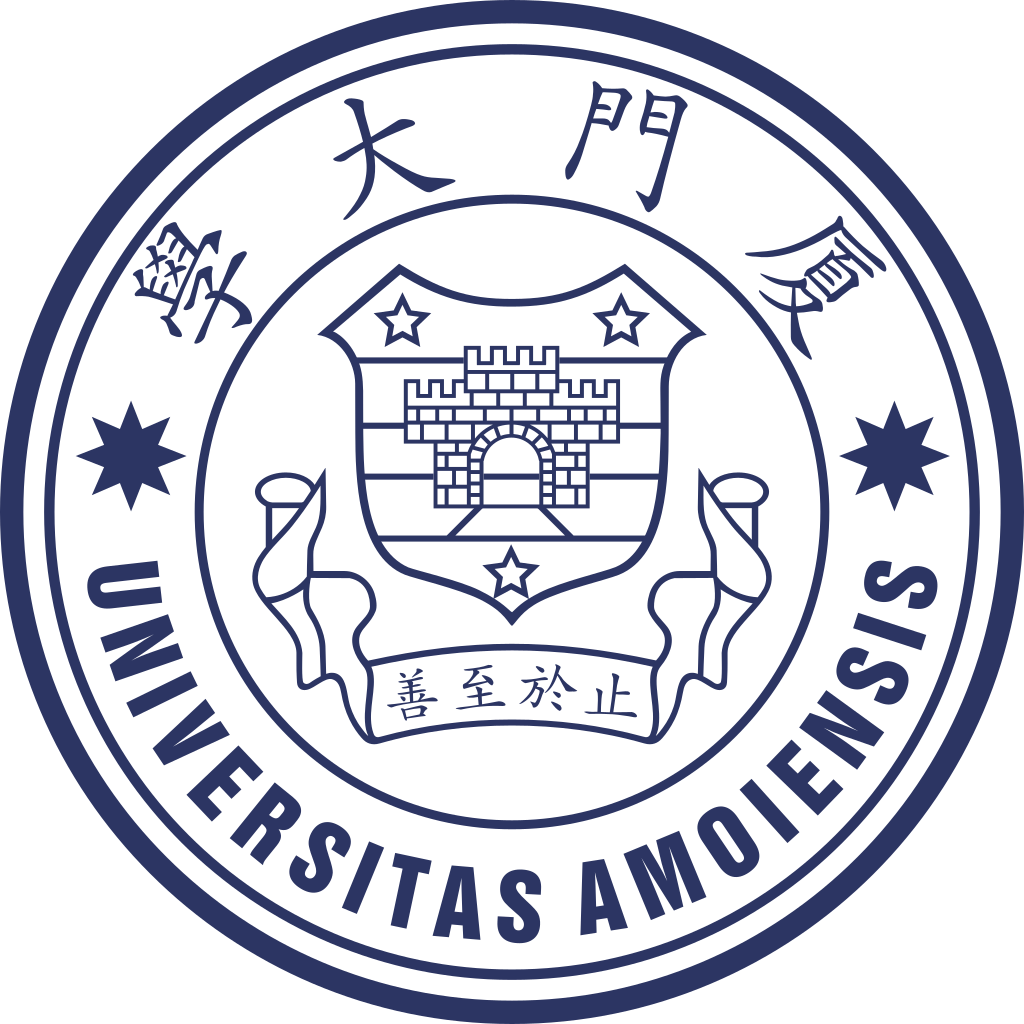}}%
  \headerlogospace{2.8mm}%
  \raisebox{-0.05\height}{\includegraphics[height=13mm,trim={9bp 14.25bp 4.5bp 5.25bp},clip]{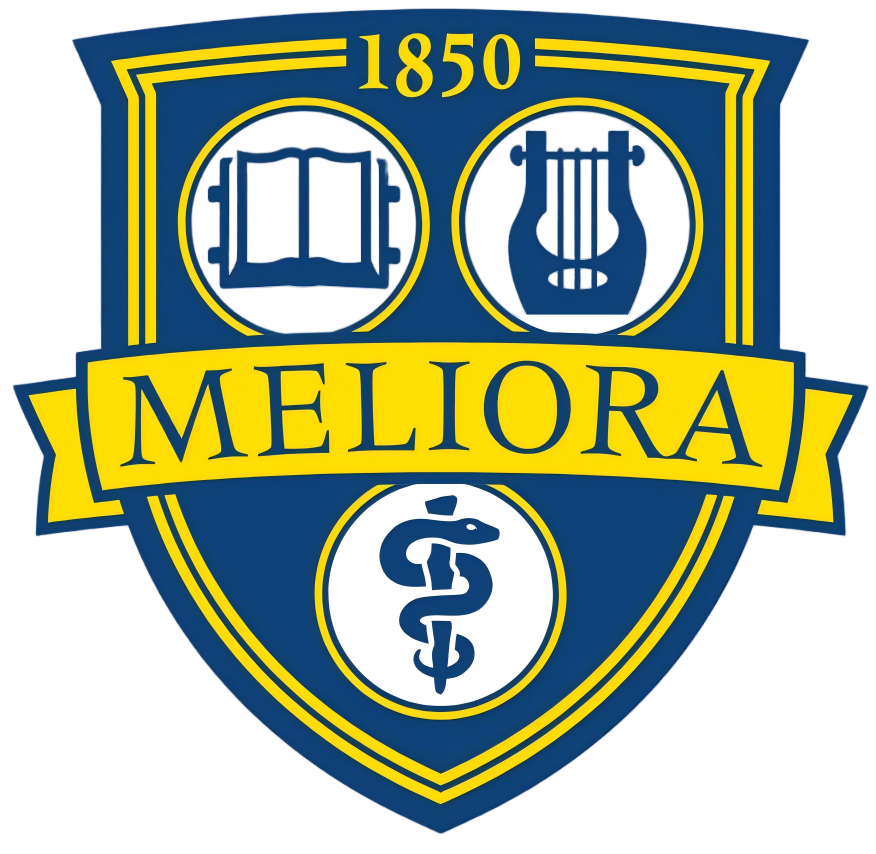}}%
}
\setrightheadericon{}
\setrunningheadericon{}
\setheadergroupname{}

\title{SCOPE: Semantic Bounded Credit Optimization for Prompt Engineering in Text-to-image Generation}
\setfrontauthors{%
  \authorrow{%
    \authorentry{Yufei Jiao}{1,*},\authorsep
    \authorentry{Jinfa Huang}{2,*},\authorsep
    \authorentry{Zeyue Xue}{4},\authorsep
    \authorentry{Tianyu Xie}{1},}%
  \authorrow{%
    \authorentry{Yuexiao Ma}{3},\authorsep
    \authorentry{Rongrong Ji}{1},\authorsep
    \authorentry{Xiawu Zheng}{1, \corremailmark},\authorsep
    \authorentry{Jiebo Luo}{2}}%
}

\setfrontaffiliations{%
  \affiliationline{%
    \authormark{1}Xiamen University \quad
    \authormark{2}University of Rochester \quad \par
    \authormark{3}Nanyang Technological University \quad
    \authormark{4}The University of Hong Kong%
  }
\vspace{2.8mm}
  {\fontsize{9.9}{11.9}\selectfont\color{black!82}%
  \ifXeTeX\omnibytesansmedium\else\sffamily\fi
  {  
  \textbf{* Equal Contributions} \par
  }}
    \vspace{-4mm}
}

\usecustomauthorlayout

\checkdata[Email]{Yufei Jiao \href{mailto:yufei@stu.xmu.edu.cn}{{\ifXeTeX\omnimonofont\else\ttfamily\fi yufei@stu.xmu.edu.cn}}, Jinfa Huang \href{mailto:jhuang90@ur.rochester.edu}{{\ifXeTeX\omnimonofont\else\ttfamily\fi jhuang90@ur.rochester.edu}}}
\checkdata[Code]{\href{https://github.com/MAC-AutoML/SCOPE}{\nolinkurl{github.com/MAC-AutoML/SCOPE}}}

\abstract{\input{content/sec/0_abstract}}

\begin{document}

\maketitle

\input{content/sec/1_intro}
\input{content/sec/2_related}
\input{content/sec/3_method}

\input{content/sec/4_experiments}
\input{content/sec/5_conclusion}

\FloatBarrier
\clearpage
\begingroup
\setlength{\bibsep}{3pt plus 1pt minus 1pt}
\hyphenpenalty=50
\exhyphenpenalty=50
\bibliographystyle{plainnat}
\bibliography{references}
\endgroup

\clearpage
\beginappendix
\input{content/sec/append}

\end{document}

%% file: content/resources/packages.tex
\usepackage{xargs}
\usepackage{todonotes}
\usepackage{tabularx}
\usetikzlibrary{calc,fit,positioning}

%% file: content/resources/math_macro.tex
\newcommand{\method}{\textbf{SCOPE}}

%% file: content/sec/0_abstract.tex
Reinforcement learning (RL) enables language models to improve text-to-image (T2I) generation by rewriting prompts for a frozen image generator. Existing methods typically apply a single signal derived from the image to individual subword tokens, although objects, counts, attributes, and relations form coherent semantic spans. We propose SCOPE, an objective based on Group Relative Policy Optimization (GRPO) that aligns update weighting with this structure. The rewriter jointly generates an enhanced prompt and its semantic partition. SCOPE pools shifts in policy log probabilities within each span to derive a shared coefficient for its token losses while retaining a single advantage for each completion. It requires neither separate span rewards nor additional image renders. Across Qwen-Image, SD3, and FLUX.1-dev, SCOPE achieves average relative gains over vanilla GRPO of 9.48\% on GenEval and 7.62\% on T2I-CompBench. Ablation studies show that both semantic boundaries and pooling within spans contribute to these gains, supporting semantic structure as a practical inductive bias for prompt engineering with RL.

%% file: content/sec/1_intro.tex
\section{Introduction}
\label{sec:introduction}

Text-to-image (T2I) generation models~\citep{hao2023optimizing,manas2024improving,rosenman2024neuroprompts,mrini2024fast,chen2025improving,yang2025self,wu2026reprompt,wu2025rewarddance,rombach2022high,ramesh2022hierarchical,saharia2022photorealistic,chefer2023attend,ho2022classifier} turn language into visual content, with prompts serving as the primary interface for controlling generation. Small wording changes can determine whether an image preserves object counts, binds colors correctly, or places objects in the intended spatial relation. As diffusion and flow generators grow larger, retraining or lightly adapting the visual decoder is often expensive and risks image quality~\citep{lipman2022flow,liu2022flow,esser2024scaling,lu2022dpm,karras2022elucidating,peebles2023scalable,podell2024sdxl,betker2023improving,chen2024pixart,cai2025hidream,gao2025seedream}. Prompt adaptation offers a cheaper alternative by improving the language interface while keeping the image generator fixed.

Prompt optimization with a frozen generator can be formulated as a reinforcement learning problem. A language policy rewrites the input prompt, the T2I model renders an image, and an evaluator assigns a scalar reward based on the requested composition. Recent methods~\citep{hao2023optimizing,manas2024improving,wu2026reprompt} exploit this feedback loop to improve T2I alignment while keeping the visual backbone fixed. GRPO converts rewards from grouped rewrites into an advantage for each completion without requiring a learned critic~\citep{shao2024deepseekmath,liu2026flow}. The objective applies this advantage across token positions without accounting for semantic span boundaries.

Compositional prompt edits to objects, attributes, counts, spatial relations, and scene context form coherent phrases that can span several subword tokens. Tokenizer boundaries reflect subword segmentation and need not coincide with these semantic units. An object description can combine an identity, a count, and an attribute, while a relation phrase links multiple objects. Words expressing each edit jointly specify a compositional constraint, even across multiple subword units. The image reward evaluates the resulting composition holistically, without direct attribution to constituent spans or tokens. This exposes a fundamental credit assignment bottleneck: uniform token-level updates fragment coherent compositional edits across arbitrary subword boundaries, yet estimating dense per-span rewards incurs prohibitive rendering overhead.

\begin{figure}[t]
  \centering
  \includegraphics[width=\linewidth]{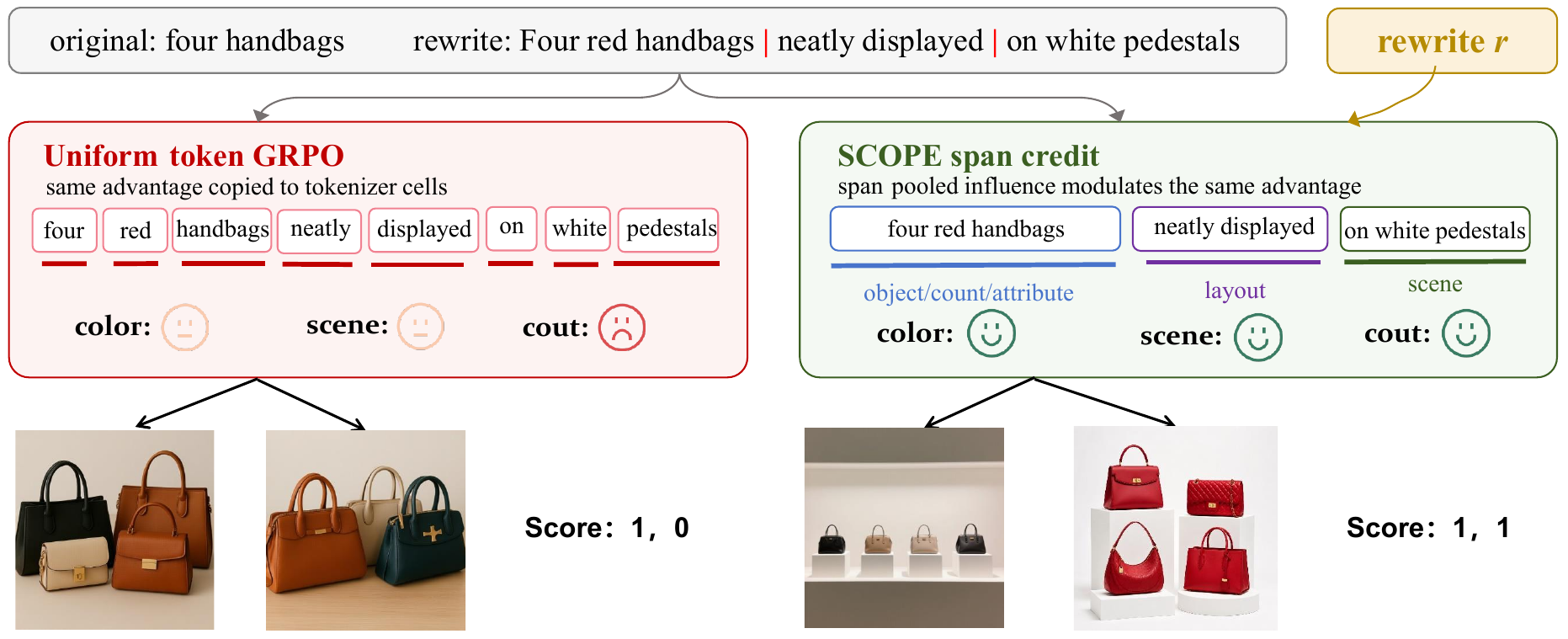}
  \caption{\textbf{Comparison of uniform token weighting in GRPO and semantic span weighting in SCOPE.} A T2I evaluator provides a single reward for the image. Standard GRPO derives an advantage for the completion and applies it across token positions. However, compositional prompt edits are naturally organized into semantic phrases whose boundaries need not align with tokenizer positions. \textbf{SCOPE} preserves the same advantage for the completion and derives a shared coefficient for each semantic span from shifts in policy log probabilities. This coefficient modulates all token updates within the span.}
  \label{fig:motivation}
\end{figure}

To bridge this gap, we introduce \textbf{S}emantic Bounded \textbf{C}redit \textbf{O}ptimization for \textbf{P}rompt \textbf{E}ngineering (\textbf{SCOPE}), a GRPO-compatible objective that modulates token updates across semantic spans. The rewriter outputs a prompt rewrite and its semantic partition, mapped onto rollout tokens. SCOPE pools shifts in token log probabilities between the current and old policies within each span to derive a detached coefficient. This coefficient reweights the contribution of the shared advantage to each token loss in the clipped GRPO surrogate. The objective retains a loss for each valid token and a single advantage for each completion. SCOPE uses the existing rendering and reward pipeline, without separate rewards or advantages for individual spans, additional image renders, or a learned critic. 
Moreover, SCOPE serves as a strict generalization of standard GRPO, seamlessly recovering uniform token updates when span modulation is neutralized ($\beta = 0$).

Across Qwen-Image, SD3, and FLUX.1-dev, SCOPE consistently outperforms vanilla GRPO on GenEval~\citep{ghosh2023geneval} and T2I-CompBench~\citep{huang2023t2i}. Ablations favor pooling policy shifts within semantic spans over weighting individual tokens, and semantic boundaries over random boundaries or fixed windows. The token weighting comparison tests the contribution of pooling over semantic units. The random and fixed window controls test the contribution of semantic boundaries beyond grouping neighboring tokens. Together, these comparisons separate semantic structure from generic reweighting. Overall, we make three contributions:

\begin{itemize}[leftmargin=1.25em,itemsep=0.2em,topsep=0.2em]
    \item We identify semantic credit fragmentation in T2I prompt rewriting with RL. A single signal per completion, derived from the image, drives updates to tokenizer positions. Compositional edits form coherent semantic spans whose boundaries need not align with the subword boundaries imposed by the tokenizer.
    \item We introduce \textbf{SCOPE}, an objective compatible with GRPO that pools policy log probability shifts within semantic spans to assign shared token weights. It retains one advantage per completion without separate span rewards or advantages, additional image renders, or a learned critic.
    \item We demonstrate that \textbf{SCOPE} consistently outperforms the vanilla GRPO baseline on GenEval and T2I-CompBench. These gains hold across three frozen T2I backbones: Qwen-Image, SD3, and FLUX.1-dev.
\end{itemize}

%% file: content/sec/2_related.tex
\section{Related Work}
\label{sec:related_work}

\paragraph{\normalfont\bfseries Credit Assignment for Sequence Generation.}
GRPO~\citep{shao2024deepseekmath} optimizes autoregressive policies without a learned critic and shares an advantage across tokens in each completion. Recent work~\citep{parthasarathi2025grpo,khandoga2026beyond,tran2025exploiting,mou2026beyond,samanta2026credit,chen2026hidden,wang2026rtmc,li2026outcome} explores finer credit assignment through eligibility traces, counterfactual evaluation, and estimation over tree structures. Segment Policy Optimization (SPO)~\citep{guo2026segment} partitions reasoning trajectories into segments and estimates distinct advantages for individual segments. This gives credit estimates a granularity between individual tokens and complete trajectories. FIPO~\citep{ma2026fipo} uses discounted future policy divergence to reweight token updates according to subsequent trajectory behavior. Our SCOPE retains a single advantage for each completion. Its contribution to token updates is modulated by shared within each semantic span, derived by pooling the policy log-probability shifts between the current and previous policies.

\paragraph{\normalfont\bfseries Prompt Optimization for T2I Generation.}
Prompt optimization adapts the textual input to elicit desired outputs from a pretrained T2I generator. Prior work~\citep{manas2024improving,rosenman2024neuroprompts,mrini2024fast,chen2025improving,xiang2025promptsculptor,kim2025reward,wu2026visualprompter,khan2025test,li2026preference,liu2026one,tang2026prism,huang2026ape} explores search and iterative refinement strategies for improving generation quality or consistency between text and images. Learned prompt adaptation trains language models to transform user inputs into prompts that the generator prefers. Promptist~\citep{hao2023optimizing} combines supervised initialization with RL driven by image rewards. PAG uses GFlowNets to promote effective and diverse prompt adaptation, while RePrompt uses GRPO and signals for image preference, alignment, and composition to train a rewriter augmented with reasoning~\citep{yun2025learning,wu2026reprompt,wang2026promptrl}. Our SCOPE addresses the optimization objective within this setting. It incorporates the semantic structure of compositional edits into token weighting while retaining a shared advantage for the complete rewrite.

\paragraph{\normalfont\bfseries Reinforcement Learning for T2I Generation.}
Reinforcement learning optimizes visual generators by treating image synthesis as a sequential decision process. Prior work~\citep{black2024training,wallace2024diffusion,wang2025pref} optimizes diffusion trajectories through policy gradients driven by image rewards or through preference optimization. Flow-GRPO~\citep{liu2026flow,lyu2025multi,ding2026treegrpo,savani2026stepwise} optimizes denoising trajectories of flow models, while Mask-GRPO~\citep{luo2026reinforcement} treats iterative masked generation as a policy with multiple steps. B$^2$-DiffuRL~\citep{hu2025towards} addresses sparse reward credit assignment within diffusion trajectories through backward progressive training and sampling over branches. Our SCOPE optimizes the prompt rewriting policy through image feedback while keeping the renderer frozen.

%% file: content/sec/3_method.tex
\section{Methodology}
\label{sec:methodology}

SCOPE organizes policy updates around semantic spans.
Given a prompt $x$, the autoregressive policy $\pi_\theta$ generates a completion $y$ with the rewrite and its span boundaries.
The cleaning operator $c(\cdot)$ extracts the \textnormal{\textless answer\textgreater} field, removes the \textnormal{\textbar} separators, and concatenates the span text.
A frozen image generator renders $I=f_\phi(c(y))$, which the evaluator scores against the cleaned prompt.
The policy therefore receives feedback for a complete rewrite, although the compositional edits within it can span several tokens.

\begin{figure}[t]
  \centering
  \includegraphics[width=\linewidth]{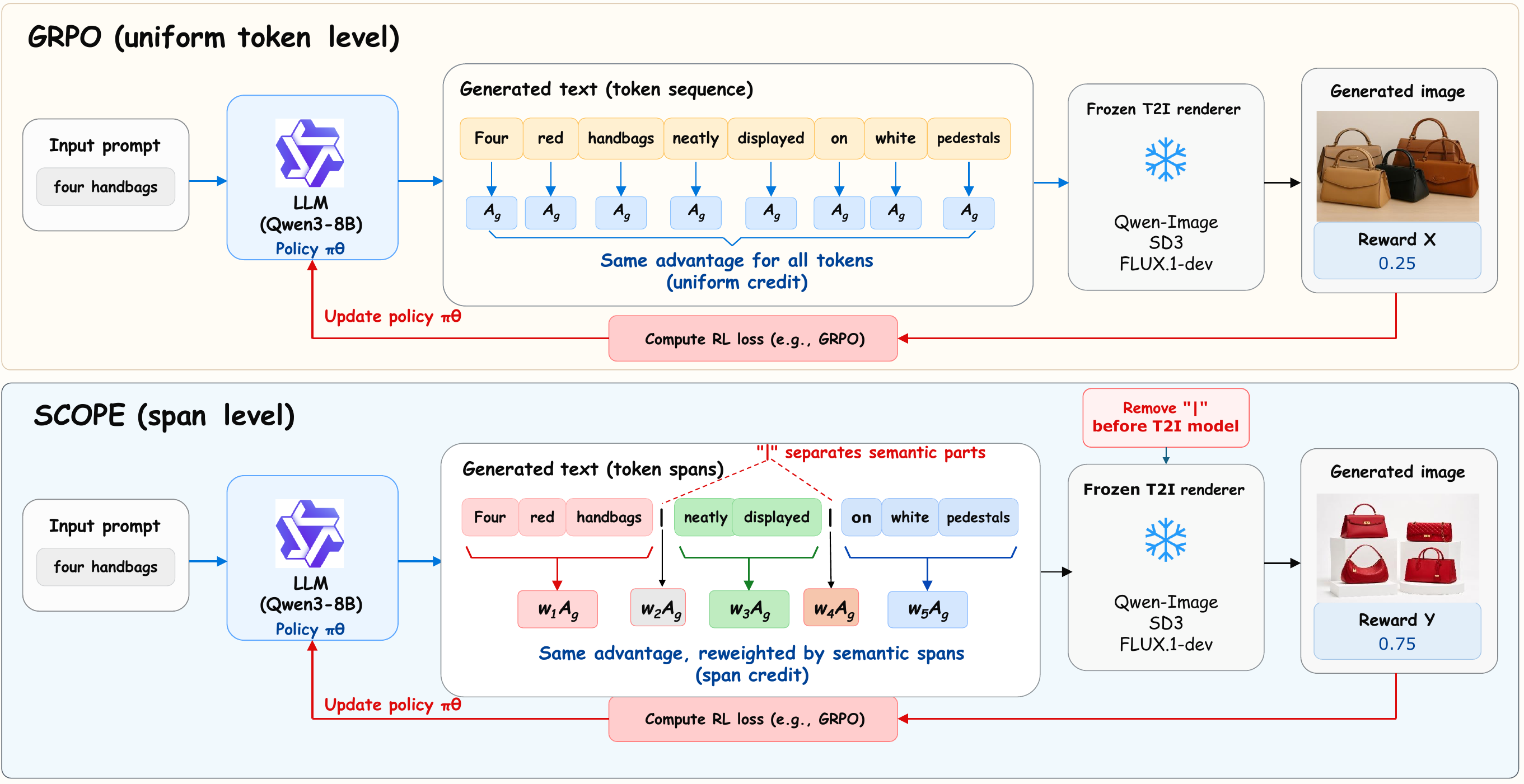}
  \caption{\textbf{Overview of SCOPE Pipeline.} The rewriter generates a prompt together with its semantic spans. After removing the span separators, the frozen renderer generates an image whose score contributes to the reward for the completion. SCOPE aligns the spans to rollout tokens and pools changes in log probability between the current and old policies. The resulting coefficients reweight the shared advantage within each span. Gradients flow through the clipped GRPO surrogate while the coefficients are detached.}
  \label{fig:scope_pipeline}
\end{figure}

Figure~\ref{fig:scope_pipeline} shows how SCOPE incorporates semantic spans into GRPO~\citep{shao2024deepseekmath}.
Grouped rollouts provide a completion advantage, while token likelihood changes determine its contribution within each semantic span.
SCOPE pools these changes over aligned spans and normalizes them into shared coefficients.
The detached coefficients reweight token losses in the clipped surrogate, with gradients flowing through the policy ratios.

\subsection{Grouped Rewards and Advantages}
\label{subsec:rollouts}

We treat prompt rewriting as a single decision whose action is the complete rewrite. The policy maximizes the expected reward over prompts drawn from the input distribution $\mathcal{D}$ and completions sampled from $\pi_\theta$:
\begin{equation}
\label{eq:rl_objective}
\max_{\theta}\;\mathbb{E}_{x\sim\mathcal{D}}\,\mathbb{E}_{y\sim\pi_\theta(\cdot\mid x)}\big[r(x,y)\big].
\end{equation}
For each minibatch prompt $x_g$, we sample $N-1$ rewrites $\{y_{g,i}\}_{i=1}^{N-1}$ from $\pi_\theta$ and append the original prompt as $y_{g,N}$.
This reference has $c(y_{g,N})=x_g$ and contributes to advantage estimation but receives no policy update.
All group members, including the reference, share a rendering seed to control renderer noise when comparing prompt edits.
The reward combines image alignment and output format:
\begin{equation}
\label{eq:reward_mix}
 r_{g,i}=w_{\mathrm{t2i}}\,r_{\mathrm{t2i}}(x_g,c(y_{g,i}))+w_{\mathrm{fmt}}\,r_{\mathrm{fmt}}(y_{g,i}).
\end{equation}
The alignment term $r_{\mathrm{t2i}}\in[0,1]$ uses the GenEval and T2I-CompBench evaluation models to score the cleaned prompt and rendered image.
The format term $r_{\mathrm{fmt}}\in[-0.25,0.25]$ evaluates the \textnormal{\textless answer\textgreater}$\cdots$\textnormal{\textless/answer\textgreater} structure and whether its spans can be extracted.
This term rewards valid structured output and penalizes malformed output independently of the image alignment score used to evaluate the rewrite.

The original prompt provides a reference for advantage estimation.
Let $b_g=r_{g,N}$ denote its reward, $\mu_g=(N-1)^{-1}\sum_{j=1}^{N-1}r_{g,j}$ the mean reward of sampled rewrites, and $\hat\sigma_g=\big((N-1)^{-1}\sum_{j=1}^{N-1}(r_{g,j}-\mu_g)^2\big)^{1/2}$ their standard deviation.
For $i<N$, we use threshold $\tau$, clipping magnitude $c$, and numerical constant $\epsilon$:
\begin{equation}
\label{eq:hybrid_advantage}
A_{g,i}=
\begin{cases}
\operatorname{clip}(r_{g,i}-b_g,\,-c,\,c), & b_g < \tau, \\[4pt]
\operatorname{clip}\!\left(\dfrac{r_{g,i}-\mu_g}{\hat{\sigma}_g+\epsilon},\,-c,\,c\right), & b_g \ge \tau,
\end{cases}
\qquad A_{g,N}=0
\end{equation}
When $b_g<\tau$, the first branch measures improvement over the original prompt.
Otherwise, the second branch standardizes rewards within the group of sampled rewrites, following GRPO~\citep{shao2024deepseekmath,liu2026flow}.

\subsection{Policy Optimization with GRPO}
\label{subsec:uniform_credit}

For completion $y_{g,i}=(y_{g,i,1},\ldots,y_{g,i,T})$, let $\mathcal{V}_{g,i}$ contain its valid token positions.
Each completion is sampled once from $\pi_\theta$, and $\pi_{\mathrm{old}}$ is the frozen pre-update snapshot, set equal to the current parameters on the first step. The probability ratio at position $t$ is
\begin{equation}
\label{eq:rho_def}
\rho_{g,i,t}=\frac{\pi_\theta(y_{g,i,t}\mid x_g,y_{g,i,<t})}{\pi_{\mathrm{old}}(y_{g,i,t}\mid x_g,y_{g,i,<t})},\qquad t\in\mathcal{V}_{g,i}
\end{equation}
GRPO applies the shared advantage $A_{g,i}$ to every valid token position through the clipped surrogate
\begin{equation}
\label{eq:grpo_surrogate}
\ell^{\mathrm{uni}}_{g,i}
=-\sum_{t\in\mathcal{V}_{g,i}}
\min\!\left(
\rho_{g,i,t}A_{g,i},\;
\operatorname{clip}(\rho_{g,i,t},1-\epsilon_\ell,1+\epsilon_h)A_{g,i}
\right).
\end{equation}
The ratios $\rho_{g,i,t}=\exp(\Delta_{g,i,t})$ vary across tokens, while the advantage and clipping margins $\epsilon_\ell,\epsilon_h$ are shared across the completion.
Each token contributes separately to the objective, even when several positions describe one object or spatial relation.
SCOPE assigns a shared loss weight to tokens within each semantic span.

\subsection{Semantic Span Credit Modulation}
\label{subsec:span_credit}

The rewrite instruction asks the policy to preserve the request and clarify subjects, counts, colors, attributes, and spatial relations.
It generates the rewrite and its boundaries in one completion, placing \textnormal{\textbar} only between adjacent nonempty spans.
Tokenizer character offsets link the extracted spans to the positions in the loss.

We extract the semantic partition from the rewrite by splitting the answer field at \textnormal{\textbar} separators and aligning the spans to rollout tokens.
Let $\mathcal{T}_{g,i,s}$ contain the valid content token positions assigned to span $s$.
These sets are nonempty and disjoint, and together they cover the answer content.
Separators and wrapper tokens lie outside the partition, so the span statistics describe the prompt passed to the renderer.

SCOPE measures the log-probability shift on the sampled completion after each update::
\begin{equation}
\label{eq:span_log_shift}
\Delta_{g,i,t}
=
\log\pi_\theta(y_{g,i,t}\mid x_g,y_{g,i,<t})
-
\log\pi_{\mathrm{old}}(y_{g,i,t}\mid x_g,y_{g,i,<t}).
\end{equation}
Averaging the log ratios $\Delta_{g,i,t}=\log\rho_{g,i,t}$ within each span removes the direct scaling with token count:
\begin{equation}
\label{eq:span_local_weight}
b_{g,i,s}
=
\frac{1}{|\mathcal{T}_{g,i,s}|}
\sum_{t\in\mathcal{T}_{g,i,s}}
\Delta_{g,i,t}.
\end{equation}
The exponential of $b_{g,i,s}$ is the geometric mean of the ratios $\rho_{g,i,t}$ within span $s$.
The sensitivity parameter $\beta$ controls the response to this aggregated signal, with clipping limiting the raw coefficient:
\begin{equation}
\label{eq:span_coefficient}
v_{g,i,s}
=
\operatorname{clip}
\left(
\exp(\beta b_{g,i,s}),
0.8,1.2
\right).
\end{equation}
The raw coefficients are positive and bounded in $[0.8,1.2]$ before normalization.
For $\beta>0$, the raw coefficient increases with this mean change until clipping saturates, and the normalized weight redistributes the current completion advantage.
The first step keeps every coefficient at one, and later steps separate spans whose mean $\Delta$ differs.
The positive coefficient preserves the sign of the completion advantage and scales the token loss contributions within that span.
We normalize by the mean weighted by valid token counts so that the average coefficient over the answer content equals one, regardless of how many tokens each span contains:
\begin{equation}
\label{eq:span_normalized_weight}
w_{g,i,s}
=
\frac{v_{g,i,s}}
{
\frac{\sum_r |\mathcal{T}_{g,i,r}|v_{g,i,r}}
{\sum_r|\mathcal{T}_{g,i,r}|}
}.
\end{equation}
Weighting this mean by span length matches the token sum in the objective and preserves ratios between span coefficients.
A common multiplicative change applied to all clipped coefficients cancels under normalization.
The mean modulated advantage over the answer content therefore remains $A_{g,i}$.
The weights redistribute loss contributions across semantic spans while the completion reward supplies their shared advantage.
The normalized coefficient is detached and applied to the advantage at each position in its span:
\begin{equation}
\label{eq:span_modulated_advantage}
\widetilde{A}_{g,i,t}
=
A_{g,i}\operatorname{stopgrad}(w_{g,i,s}),
\quad
t\in\mathcal{T}_{g,i,s}
\end{equation}
In this surrogate, detaching aggregation, clipping, and normalization restricts gradients to $\rho_{g,i,t}$.
Valid wrapper and separator positions retain the original advantage, $\widetilde{A}_{g,i,t}=A_{g,i}$, in the resulting loss:
\begin{equation}
\label{eq:scope_surrogate}
\ell^{\mathrm{scope}}_{g,i}
=-\sum_{t\in\mathcal{V}_{g,i}}
\min\!\left(
\rho_{g,i,t}\widetilde{A}_{g,i,t},\;
\operatorname{clip}(\rho_{g,i,t},1-\epsilon_\ell,1+\epsilon_h)\widetilde{A}_{g,i,t}
\right).
\end{equation}
The positive, detached coefficients rescale both branches of each clipped token loss equally.
They change how much of the loss each span receives without changing the clipping decision at a fixed $\rho_{g,i,t}$.

The policy objective averages the losses over the $N-1$ sampled rewrites for each prompt in minibatch $\mathcal{B}$:
\begin{equation}
\label{eq:outer_loss}
\mathcal{L}_{\mathrm{pol}}=
\frac{1}{|\mathcal{B}|}\sum_{g\in\mathcal{B}}\frac{1}{N-1}\sum_{i=1}^{N-1}\ell^{\mathrm{scope}}_{g,i}.
\end{equation}
An optional KL penalty regularizes the policy against a frozen reference~\citep{liu2026flow,xue2025dancegrpo}.
Span coefficients add one forward through $\pi_{\mathrm{old}}$ and require no additional model calls or image renders. Because the rewriter jointly learns structural formatting via $r_{\text{fmt}}$, it achieves over 99.2\% exact span-to-token alignment across all training rollouts (detailed in Appendix~\ref{sec:app_fallbacks}). For the negligible fraction of unaligned rollouts ($<0.8\%$), defaulting the modulation weight to $w_{g,i,s}=1$ guarantees uninterrupted optimization stability.

%% file: content/sec/4_experiments.tex
  \begin{table}[!t]
  \centering
  \captionsetup{font=footnotesize}
  \caption{\textbf{SCOPE achieves the highest overall GenEval score on each frozen backbone.} RePrompt$^{\dagger}$ is our Qwen-Image reproduction, while its SD3 and FLUX.1-dev scores follow \citet{wu2026reprompt}. Single and Two denote one and two objects. Cnt., Clr., Pos., and Attr. denote counting, color, position, and attribute binding. Bold marks the best score per metric and backbone.}

  \label{tab:geneval}
  \normalsize
  \setlength{\tabcolsep}{2.5pt}
  \renewcommand{\arraystretch}{1.04}
  \begin{tabularx}{\linewidth}{@{}ll*{7}{>{\centering\arraybackslash}X}@{}}
  \toprule
  \textbf{Backbone} & \textbf{Method}
  & \textbf{Single} & \textbf{Two} & \textbf{Cnt.} & \textbf{Clr.} & \textbf{Pos.} & \textbf{Attr.}
  & \textbf{Overall}$\uparrow$ \\
  \midrule
  \multirow{5}{*}{Qwen-Image}
  & Baseline                                & 0.99 & 0.95 & 0.87 & 0.89 & 0.79 & 0.75 & 0.87 \\
  & Z-Image rewriter        & 0.95 & 0.89 & 0.69 & 0.82 & 0.51 & 0.73 & 0.76 \\
  & GRPO                                    & 0.99 & 0.97 & 0.90 & \textbf{0.94} & 0.87 & 0.80 & 0.91 \\
  & RePrompt$^{\dagger}$~\citep{wu2026reprompt}   & 0.99 & 0.95 & 0.83 & 0.85 & 0.69 & 0.66 & 0.83 \\
  \rowcolor{cyan!10}
  & \textbf{SCOPE} (Ours)               & \textbf{1.00} & \textbf{0.99} & \textbf{0.94} & 0.93 & \textbf{0.95} & \textbf{0.88} & \textbf{0.95} \\
  \cmidrule(lr){1-9}
  \multirow{5}{*}{SD3}
  & Baseline                                & 0.98 & 0.78 & 0.44 & 0.83 & 0.23 & 0.50 & 0.63 \\
  & Z-Image rewriter        & 0.58 & 0.54 & 0.37 & 0.42 & 0.25 & 0.37 & 0.42 \\
  & GRPO                                    & 0.98 & 0.87 & 0.58 & 0.85 & 0.57 & 0.60 & 0.74 \\
  & RePrompt~\citep{wu2026reprompt}               & 0.99 & 0.86 & 0.60 & 0.86 & 0.59 & 0.60 & 0.75 \\
  \rowcolor{cyan!10}
  & \textbf{SCOPE} (Ours)               & \textbf{1.00} & \textbf{0.90} & \textbf{0.79} & \textbf{0.89} & \textbf{0.83} & \textbf{0.63} & \textbf{0.84} \\
  \cmidrule(lr){1-9}
  \multirow{5}{*}{FLUX.1-dev}
  & Baseline                                & 0.99 & 0.80 & 0.72 & 0.80 & 0.16 & 0.46 & 0.65 \\
  & Z-Image rewriter        & 0.53 & 0.50 & 0.56 & 0.42 & 0.30 & 0.36 & 0.44 \\
  & GRPO                                    & 0.98 & 0.85 & 0.72 & 0.82 & 0.62 & 0.56 & 0.76 \\
  & RePrompt~\citep{wu2026reprompt}               & 0.98 & 0.87 & 0.77 & 0.85 & 0.62 & 0.49 & 0.76 \\
  \rowcolor{cyan!10}
  & \textbf{SCOPE} (Ours)               & \textbf{1.00} & \textbf{0.89} & \textbf{0.83} & \textbf{0.89} & \textbf{0.84} & \textbf{0.60} & \textbf{0.84} \\
  \bottomrule
  \end{tabularx}
  \end{table}

\section{Experiments}
\label{sec:experiments}

\subsection{Experimental Setup}
\label{subsec:exp_setup}

\noindent\textbf{Benchmarks.}
\label{sec:app_reward_data}
We evaluate compositional prompt adherence on GenEval~\citep{ghosh2023geneval} and T2I-CompBench~\citep{huang2023t2i}.
GenEval tests object presence, counting, color, position, and attribute binding.
T2I-CompBench covers attribute binding, object relationships, and complex composition.
GenEval training uses 10,000 of the 50,000 prompts organized by Flow-GRPO~\citep{liu2026flow}, while T2I-CompBench uses the 4200 prompts in its official training split.
Both native benchmark evaluation sets are disjoint from the training data.

\noindent\textbf{Baselines.}
GRPO with uniform token weights is our primary baseline and shares SCOPE's reward pipeline and training setup.
Other baselines render original prompts directly or use the Z-Image rewriter~\citep{cai2025z} without further training.
We reproduce the published RePrompt~\citep{wu2026reprompt} training and evaluation recipe on Qwen-Image with the same reward clients, rendering seeds, and benchmark harness as our runs.
For SD3 and FLUX.1-dev, we use the published RePrompt scores.
Appendix~\ref{sec:app_t2i_projection} documents the reproduction and score sources.

\noindent\textbf{Implementation details.}
We train Qwen3-8B~\citep{yang2025qwen3} as the prompt rewriter with frozen Qwen-Image~\citep{wu2025qwen}, SD3~\cite{esser2024scaling}, and FLUX.1-dev~\cite{flux2024} generators.
We use a learning rate of $10^{-6}$, a KL coefficient of $0.01$, and a maximum completion length of 512 tokens.
Each group contains $N=8$ prompt members, including the original prompt and sampled rewrites.
All members share one rendering seed to control variation from image generation within the group.
Appendix~\ref{sec:app_hyper} includes optimization configuration and span weight normalization.

\noindent\textbf{Evaluation protocol.}
Each backbone uses fixed resolution, denoising steps, and classifier-free guidance during training and evaluation (Appendix~\ref{sec:app_t2i_gen}).
We average scores from four images per GenEval prompt and ten per T2I-CompBench prompt with distinct randomly selected seeds.
Only training groups share seeds.

\subsection{Quantitative Analysis}
\label{subsec:geneval_main}

\noindent\textbf{GenEval.} SCOPE achieves the highest overall GenEval score across all three frozen backbones (Table~\ref{tab:geneval}), boosting overall performance over vanilla GRPO from $0.91$ to $0.95$ on Qwen-Image ($+4.4\%$ relative), $0.74$ to $0.84$ on SD3 ($+13.5\%$ relative), and $0.76$ to $0.84$ on FLUX.1-dev ($+10.5\%$ relative). Across the three backbones, SCOPE achieves strict improvements over GRPO in $17$ out of $18$ category benchmarks while maintaining near-saturated color accuracy on Qwen-Image ($0.93$), trading a marginal $0.01$ fluctuation for substantial gains in complex spatial positioning ($+0.08$) and attribute binding ($+0.08$). Moreover, the most pronounced improvements appear in object counting and spatial positioning, two persistent bottlenecks for frozen T2I generators. On SD3, SCOPE substantially outperforms GRPO in both Counting ($0.58 \rightarrow 0.79$, $+36.2\%$) and Position ($0.57 \rightarrow 0.83$, $+45.6\%$). On FLUX.1-dev, SCOPE lifts Counting by $+0.11$ and raises Position accuracy from the unadapted baseline of $0.16$ and GRPO's $0.62$ to $0.84$, representing a more than $5\times$ gain over the base generator. These improvements also extend to multi-object synthesis and attribute binding, achieving $0.99$ two-object accuracy and $0.88$ attribute binding accuracy on Qwen-Image.

\FloatBarrier

\noindent\textbf{T2I-CompBench.}
\label{subsec:t2icomp}
SCOPE exceeds GRPO in all six reported categories on each frozen backbone (Table~\ref{tab:benchmark_summary_across_backbones}).
Average scores improve by 6.7\%, 8.0\%, and 8.2\% on Qwen-Image, SD3, and FLUX.1-dev, respectively.
The gains cover appearance attributes, spatial relations, numeracy, and complex composition. Notably, on strong backbones like Qwen-Image, uniform token-level RL (e.g., GRPO and RePrompt) suffers from compositional degradation, falling below the unadapted baseline in overall T2I-CompBench score ($0.6247$ and $0.6246$ vs.\ $0.6280$) as well as Texture, Numeracy, and Complex composition, as uniform token updates inadvertently distort fine-grained descriptive phrases during exploration. SCOPE completely resolves this negative transfer, reaching $0.6665$ overall and surpassing both the unadapted baseline and GRPO across all individual categories. Beyond achieving the highest color, shape, and texture scores on SD3 and FLUX.1-dev, SCOPE substantially improves spatial relations and numeracy over GRPO on both backbones ($+0.0815$ and $+0.0686$ on SD3; $+0.0491$ and $+0.0658$ on FLUX.1-dev). In Complex composition, it lifts Qwen-Image from $0.3921$ to $0.4473$ with consistent gains on SD3 and FLUX.1-dev, demonstrating robust, precise, and fine-grained control over both individual attributes and multi-constraint prompts.

\begin{table}[!t]
\centering
\captionsetup{font=footnotesize}
\normalsize
\setlength{\tabcolsep}{2.5pt}
\renewcommand{\arraystretch}{1.04}
\caption{\textbf{SCOPE exceeds GRPO in every T2I-CompBench category on all three backbones.} RePrompt$^\dag$ is our Qwen-Image reproduction. Other RePrompt scores follow \citet{wu2026reprompt}. Bold marks the best per metric and backbone.}
\label{tab:benchmark_summary_across_backbones}
\begin{tabularx}{\linewidth}{@{}ll*{4}{>{\hsize=.96\hsize\linewidth=\hsize\centering\arraybackslash}X}>{\hsize=1.2\hsize\linewidth=\hsize\centering\arraybackslash}X>{\centering\arraybackslash}X>{\hsize=.96\hsize\linewidth=\hsize\centering\arraybackslash}X@{}}
\toprule
\textbf{Backbone} & \textbf{Method} & \textbf{Color} & \textbf{Shape} & \textbf{Texture} & \textbf{Spatial} & \textbf{Numeracy} & \textbf{Complex} & \textbf{Avg} \\
\midrule
\multirow{5}{*}{Qwen-Image}
& Baseline & 0.8352 & 0.5965 & 0.7598 & 0.4452 & 0.7340 & 0.3971 & 0.6280 \\
& Z-Image rewriter & 0.8324 & 0.5930 & 0.7285 & 0.3880 & 0.6963 & 0.3922 & 0.6051 \\
& GRPO & 0.8344 & 0.6042 & 0.7366 & 0.4487 & 0.7321 & 0.3921 & 0.6247 \\
& RePrompt~\citep{wu2026reprompt}$^\dag$ & 0.8337 & 0.6110 & 0.7546 & 0.4268 & 0.7332 & 0.3882 & 0.6246 \\
\rowcolor{cyan!10}  & \textbf{SCOPE} (Ours) & \textbf{0.8633} & \textbf{0.6273} & \textbf{0.7857} & \textbf{0.5043} & \textbf{0.7713} & \textbf{0.4473} & \textbf{0.6665} \\
\midrule
\multirow{5}{*}{SD3}
& Baseline & 0.7886 & 0.5684 & 0.7267 & 0.2299 & 0.5761 & 0.3754 & 0.5442 \\
& Z-Image rewriter & 0.7933 & 0.5846 & 0.7141 & 0.2977 & 0.6286 & 0.3876 & 0.5677 \\
& GRPO & 0.7894 & 0.5742 & 0.7196 & 0.3312 & 0.6277 & 0.3802 & 0.5704 \\
& RePrompt~\citep{wu2026reprompt} & 0.7866 & 0.5891 & 0.7184 & 0.3315 & 0.6289 & 0.3744 & 0.5715 \\
\rowcolor{cyan!10}  & \textbf{SCOPE} (Ours) & \textbf{0.8213} & \textbf{0.6147} & \textbf{0.7523} & \textbf{0.4127} & \textbf{0.6963} & \textbf{0.3990} & \textbf{0.6161} \\
\midrule
\multirow{5}{*}{FLUX.1-dev}
& Baseline & 0.7414 & 0.4764 & 0.5937 & 0.2505 & 0.6099 & 0.3712 & 0.5072 \\
& Z-Image rewriter & 0.7624 & 0.5363 & 0.6396 & 0.3162 & 0.6492 & 0.3817 & 0.5476 \\
& GRPO & 0.7488 & 0.5289 & 0.6143 & 0.3742 & 0.6489 & 0.3742 & 0.5482 \\
& RePrompt~\citep{wu2026reprompt} & 0.7501 & 0.5276 & 0.6515 & 0.3301 & 0.6499 & 0.3721 & 0.5469 \\
\rowcolor{cyan!10} & \textbf{SCOPE} (Ours) & \textbf{0.7967} & \textbf{0.5677} & \textbf{0.6637} & \textbf{0.4233} & \textbf{0.7147} & \textbf{0.3913} & \textbf{0.5929} \\
\bottomrule
\end{tabularx}
\end{table}

\begin{figure}[!t]
\centering
\includegraphics[width=0.8\textwidth]{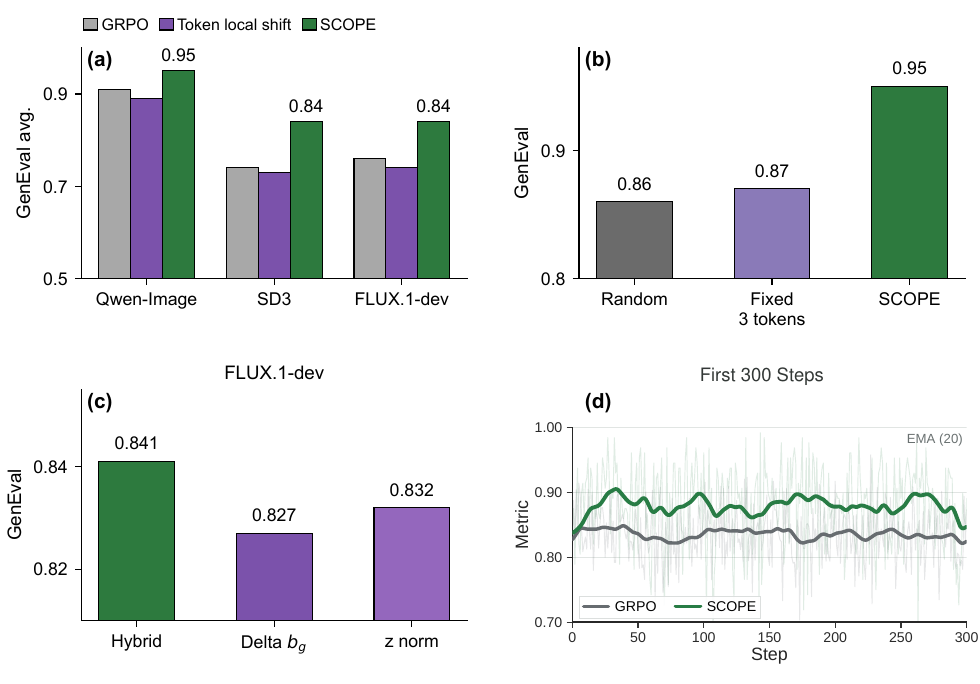}
\caption{\textbf{Ablation and mechanistic diagnostics.} (a) GenEval averages across backbones: SCOPE consistently outperforms uniform GRPO, whereas unpooled token-level shifts degrade performance. (b) Semantic boundaries substantially outperform random partitions and fixed $3$-token windows on Qwen-Image. (c) SCOPE remains robust across advantage formulations on FLUX.1-dev, with all variants surpassing vanilla GRPO ($0.760$) and the hybrid rule achieving the peak score. (d) Training reward dynamics on Qwen-Image: SCOPE breaks through GRPO's early saturation plateau to achieve higher asymptotic reward.}
\label{fig:ablation}
\end{figure}

\subsection{Ablation and Mechanistic Analysis}

\noindent\textbf{Effect of credit weighting.} Figure~\ref{fig:ablation}(a) compares SCOPE's span-pooled weighting against uniform GRPO and unpooled token-level likelihood shifts. Crucially, applying likelihood shifts at individual tokens degrades performance below standard GRPO across all backbones. This confirms our core hypothesis: unpooled token-level modulation injects high-variance subword noise that disrupts coherent multi-token phrases, whereas pooling shifts within semantic spans stabilizes credit allocation. Furthermore, a \textit{Token local shift} control, which retains semantic boundaries while applying uniform GRPO updates, yields only $0.90$, $0.75$, and $0.76$ across backbones compared to $0.95$, $0.84$, and $0.84$ for SCOPE. Thus, explicit credit modulation across spans (as opposed to merely structured formatting) drives the performance gain.

\noindent\textbf{Effect of span boundaries.} 
In Figure~\ref{fig:ablation}(b), replacing semantic partitions with random boundaries ($0.86$) or fixed $3$-token windows ($0.87$) causes a severe $8\text{--}9\%$ drop relative to SCOPE ($0.95$) on Qwen-Image. Arbitrary groupings sever coherent compositional phrases across window borders, proving that linguistic boundary alignment, rather than generic multi-token smoothing, which is essential for effective credit assignment.

\noindent\textbf{Effect of the advantage rule.} 
Figure~\ref{fig:ablation}(c) evaluates SCOPE under three completion-level advantage estimators on FLUX.1-dev: pure baseline subtraction ($\Delta b_g$, $0.827$), standard group normalization ($z$-norm, $0.832$), and our hybrid threshold rule ($0.841$). First, all three variants substantially outperform uniform GRPO ($0.760$), confirming that SCOPE's gains stem fundamentally from semantic span modulation. Second, the hybrid rule yields an additional $+0.009$ boost by anchoring updates to the original prompt when initial quality is low ($b_g < \tau$) and switching to fine-grained group discrimination once generation competence is established.

\noindent\textbf{Training dynamics.} 
Figure~\ref{fig:ablation}(d) reports the Qwen-Image training metric over the first 300 steps, with curves smoothed by a 20-step exponential moving average. The smoothed trajectory of uniform GRPO remains lower and varies only mildly, at approximately $0.83$--$0.85$. The smoothed trajectory of SCOPE lies above that of GRPO throughout the observed window and falls mainly in the range of approximately $0.86$--$0.90$, whereas its unsmoothed trace exhibits substantially larger fluctuations.

\subsection{Qualitative Analysis}
\label{subsec:qualitative_main}
\begin{figure}[t]
\centering
\footnotesize
\sffamily
\setlength{\fboxsep}{0pt}%
\begin{adjustbox}{max width=0.75\linewidth, max height=0.58\textheight, center}
\begin{tikzpicture}[x=1cm,y=1cm]
\tikzset{%
  cell/.style={draw=black!55, fill=gray!14, minimum width=2.52cm, minimum height=2.52cm, align=center, inner sep=1.5pt, font=\footnotesize},
  hdr/.style={minimum width=2.52cm, align=center, font=\bfseries\footnotesize},
}
\def\dx{2.62}
\pgfmathsetmacro{\rowgapcaption}{2.98}
\pgfmathsetmacro{\yone}{-1.88}
\node[hdr] (H1) at ({0.5*\dx},0.20) {User prompt};
\node[hdr] (HG) at ({1.5*\dx},0.20) {GRPO};
\node[hdr, text width=2.52cm] (H2) at ({2.5*\dx},0.20) {gpt-image~2};
\node[hdr] (H3) at ({3.5*\dx},0.20) {Rewrite / rationale};
\node[hdr] (H4) at ({4.5*\dx},0.20) {\textbf{SCOPE} (Ours)};
\pgfmathsetmacro{\y}{\yone}
\node[draw=black!55, fill=white, minimum width=2.52cm, minimum height=2.52cm, inner sep=0.5pt, align=center] (R1C1) at ({0.5*\dx},\y) {\includegraphics[width=2.44cm,height=2.44cm,keepaspectratio]{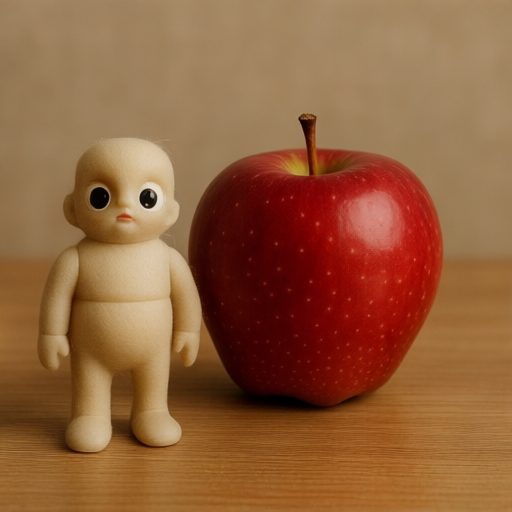}};
\node[draw=black!55, fill=white, minimum width=2.52cm, minimum height=2.52cm, inner sep=0.5pt, align=center] (R1G) at ({1.5*\dx},\y) {\includegraphics[width=2.44cm,height=2.44cm,keepaspectratio]{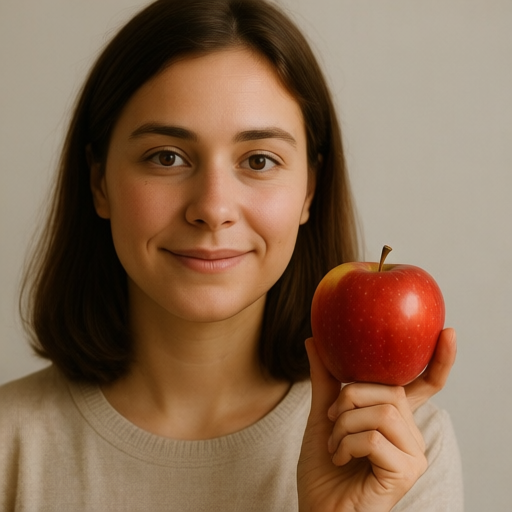}};
\node[draw=black!55, fill=white, minimum width=2.52cm, minimum height=2.52cm, inner sep=0.5pt, align=center] (R1C2) at ({2.5*\dx},\y) {\includegraphics[width=2.44cm,height=2.44cm,keepaspectratio]{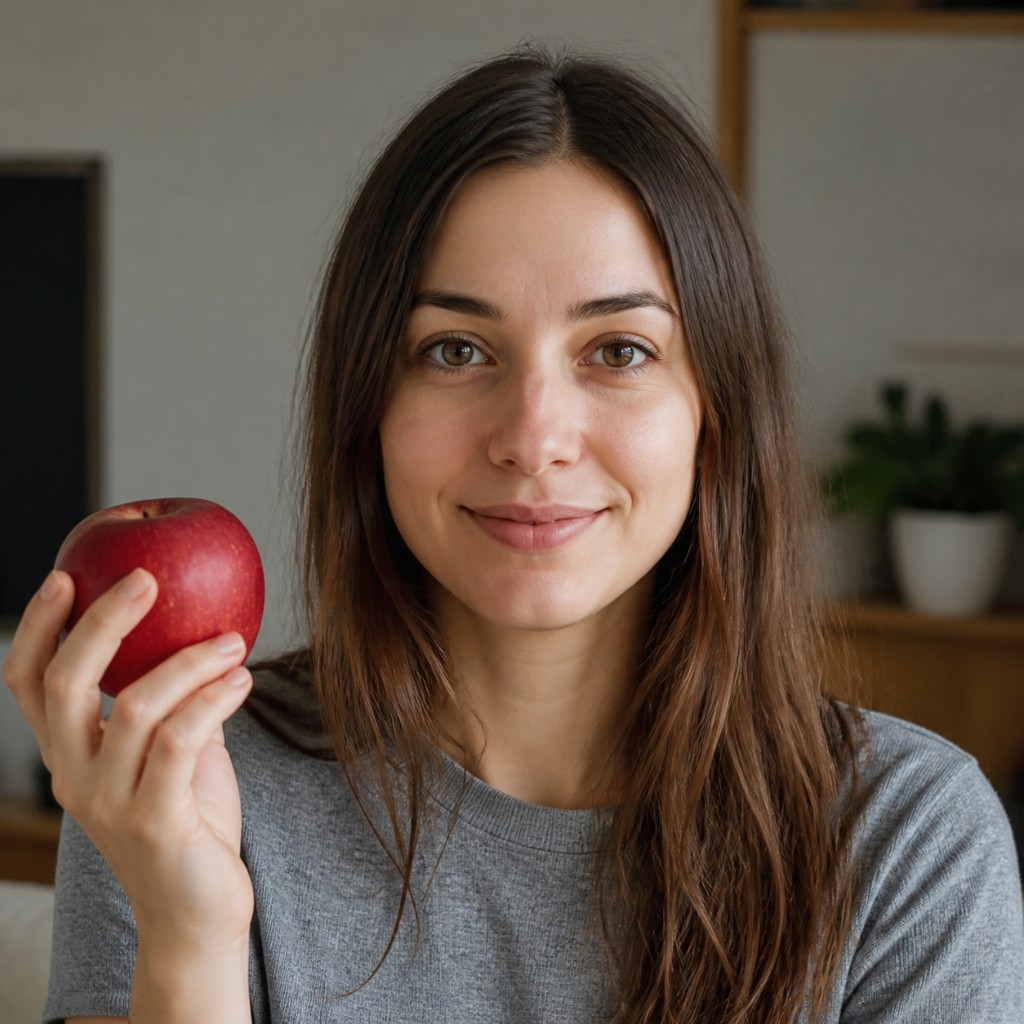}};
\node[draw=black!55, fill=white, minimum width=2.52cm, minimum height=2.52cm, inner sep=1pt, text width=2.30cm, align=left, font=\tiny] (R1C3) at ({3.5*\dx},\y) {A photo of a person holding an apple: person on the left, apple on the right; facing the camera with natural light; vibrant red apple\,\ldots\ simple neutral background, shallow DoF.};
\node[draw=black!55, fill=white, minimum width=2.52cm, minimum height=2.52cm, inner sep=0.5pt, align=center] (R1C4) at ({4.5*\dx},\y) {\includegraphics[width=2.44cm,height=2.44cm,keepaspectratio]{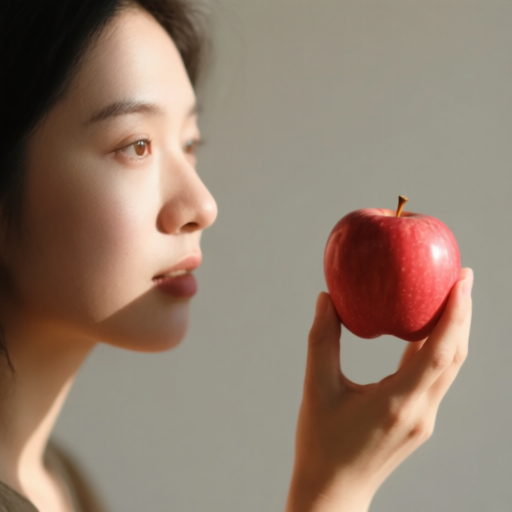}};
\node[anchor=north west, align=left, font=\scriptsize\rmfamily, inner sep=0pt] at ($(R1C1.south west)+(0,-0.04)$) {\mbox{Prompt: ``a photo of a person and an apple''}};
\pgfmathsetmacro{\yrowtwo}{\yone-\rowgapcaption}
\node[draw=black!55, fill=white, minimum width=2.52cm, minimum height=2.52cm, inner sep=0.5pt, align=center] (R2C1) at ({0.5*\dx},\yrowtwo) {\includegraphics[width=2.44cm,height=2.44cm,keepaspectratio]{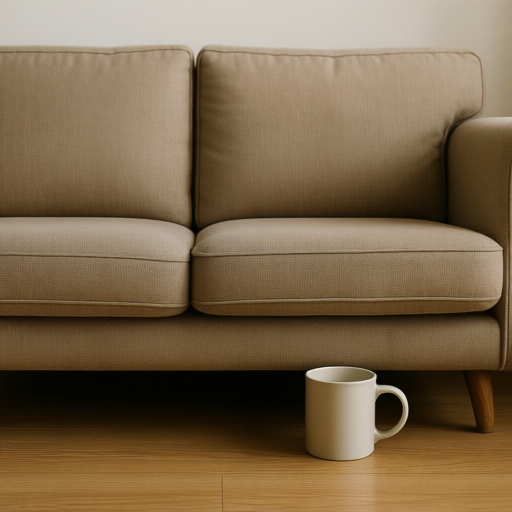}};
\node[draw=black!55, fill=white, minimum width=2.52cm, minimum height=2.52cm, inner sep=0.5pt, align=center] (R2G) at ({1.5*\dx},\yrowtwo) {\includegraphics[width=2.44cm,height=2.44cm,keepaspectratio]{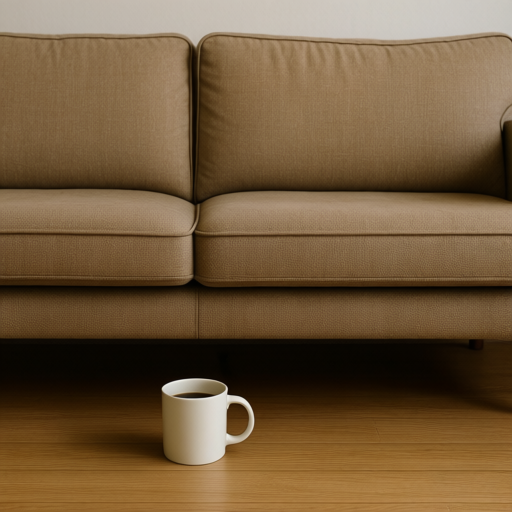}};
\node[draw=black!55, fill=white, minimum width=2.52cm, minimum height=2.52cm, inner sep=0.5pt, align=center] (R2C2) at ({2.5*\dx},\yrowtwo) {\includegraphics[width=2.44cm,height=2.44cm,keepaspectratio]{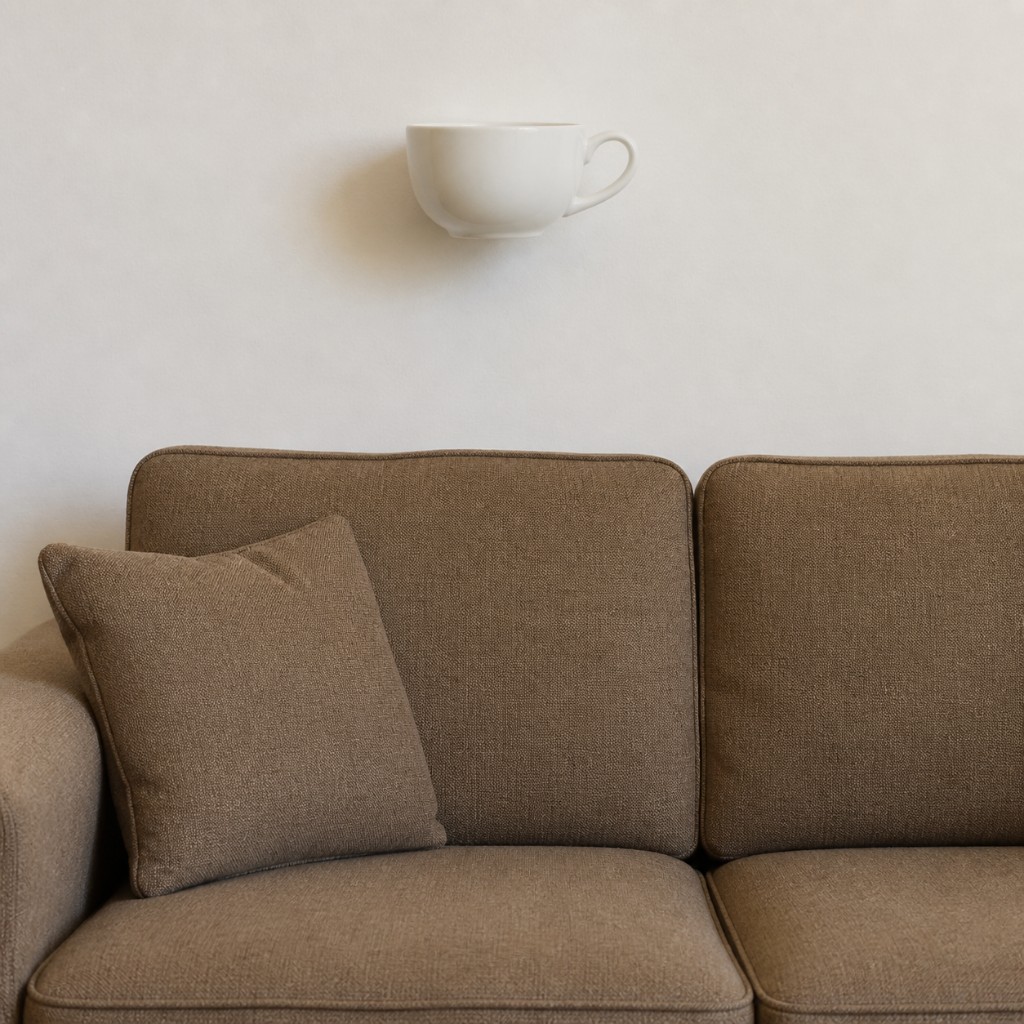}};
\node[draw=black!55, fill=white, minimum width=2.52cm, minimum height=2.52cm, inner sep=1pt, text width=2.30cm, align=left, font=\tiny] (R2C3) at ({3.5*\dx},\yrowtwo) {A photo showing a couch below a cup: cup on a surface above the couch; couch lower, cup upper, vertical composition. Natural lighting\,\ldots\ minimal atmosphere, spatial relationship emphasized.};
\node[draw=black!55, fill=white, minimum width=2.52cm, minimum height=2.52cm, inner sep=0.5pt, align=center] (R2C4) at ({4.5*\dx},\yrowtwo) {\includegraphics[width=2.44cm,height=2.44cm,keepaspectratio]{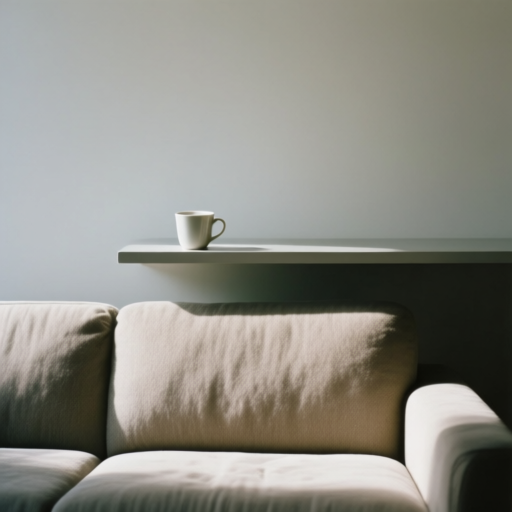}};
\node[anchor=north west, align=left, font=\scriptsize\rmfamily, inner sep=0pt] at ($(R2C1.south west)+(0,-0.04)$) {\mbox{Prompt: ``a photo of a couch below a cup''}};
\pgfmathsetmacro{\yrowthree}{\yrowtwo-\rowgapcaption}
\node[draw=black!55, fill=white, minimum width=2.52cm, minimum height=2.52cm, inner sep=0.5pt, align=center] (R3C1) at ({0.5*\dx},\yrowthree) {\includegraphics[width=2.44cm,height=2.44cm,keepaspectratio]{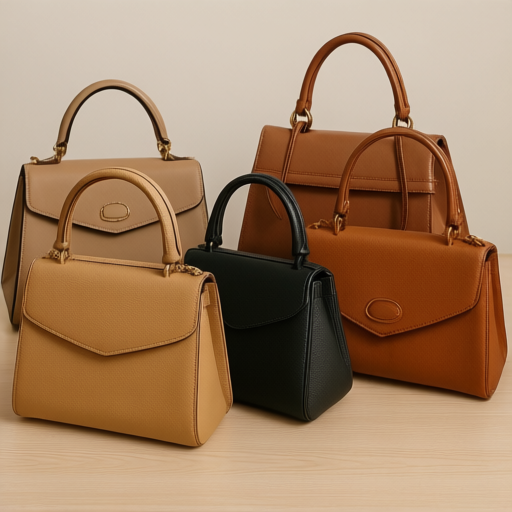}};
\node[draw=black!55, fill=white, minimum width=2.52cm, minimum height=2.52cm, inner sep=0.5pt, align=center] (R3G) at ({1.5*\dx},\yrowthree) {\includegraphics[width=2.44cm,height=2.44cm,keepaspectratio]{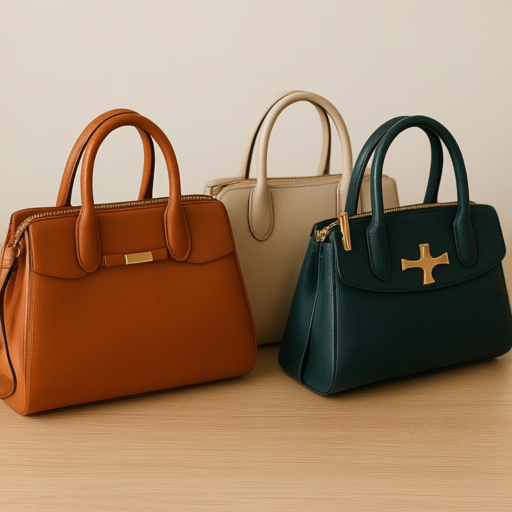}};
\node[draw=black!55, fill=white, minimum width=2.52cm, minimum height=2.52cm, inner sep=0.5pt, align=center] (R3C2) at ({2.5*\dx},\yrowthree) {\includegraphics[width=2.44cm,height=2.44cm,keepaspectratio]{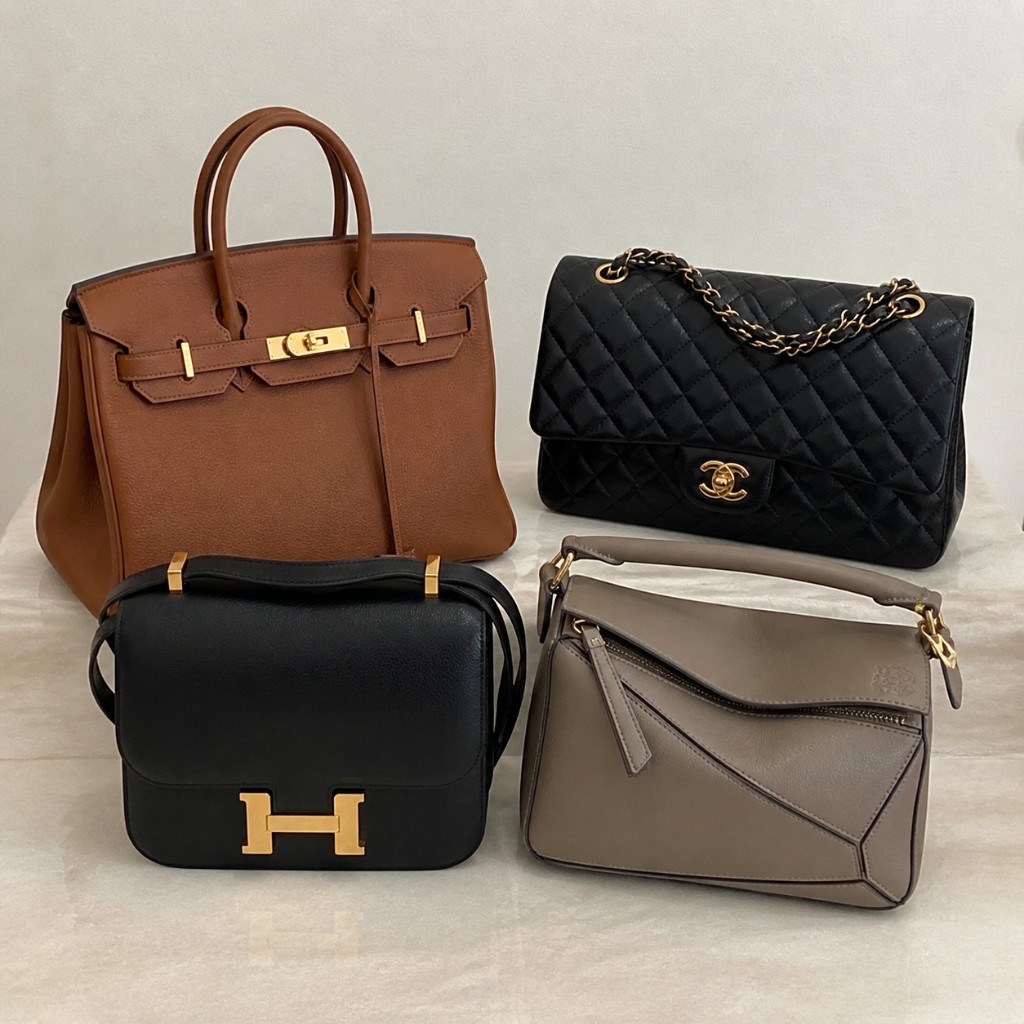}};
\node[draw=black!55, fill=white, minimum width=2.52cm, minimum height=2.52cm, inner sep=1pt, text width=2.30cm, align=left, font=\tiny] (R3C3) at ({3.5*\dx},\yrowthree) {A photo of four handbags in a neat row across the center, each on a white pedestal; equal spacing on one horizontal plane. Soft even lighting\,\ldots\ clean neutral background for clarity and texture.};
\node[draw=black!55, fill=white, minimum width=2.52cm, minimum height=2.52cm, inner sep=0.5pt, align=center] (R3C4) at ({4.5*\dx},\yrowthree) {\includegraphics[width=2.44cm,height=2.44cm,keepaspectratio]{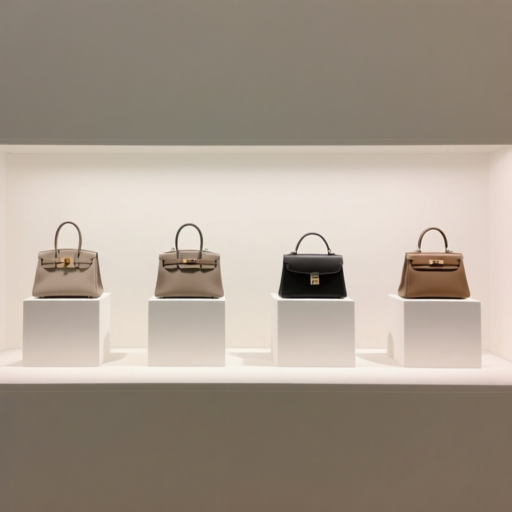}};
\node[anchor=north west, align=left, font=\scriptsize\rmfamily, inner sep=0pt] at ($(R3C1.south west)+(0,-0.04)$) {\mbox{Prompt: ``a photo of four handbags''}};
\pgfmathsetmacro{\yrowfour}{\yrowthree-\rowgapcaption}
\node[draw=black!55, fill=white, minimum width=2.52cm, minimum height=2.52cm, inner sep=0.5pt, align=center] (R4C1) at ({0.5*\dx},\yrowfour) {\includegraphics[width=2.44cm,height=2.44cm,keepaspectratio]{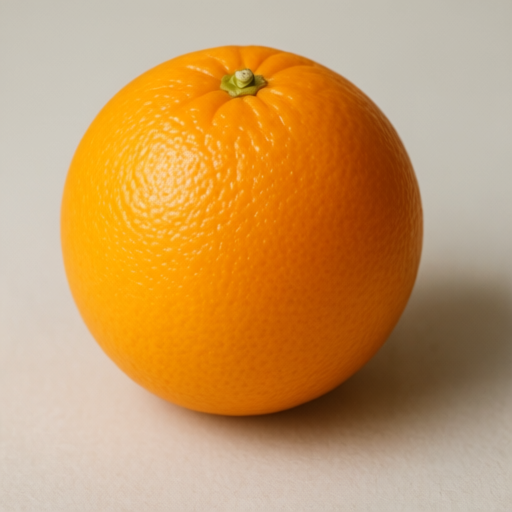}};
\node[draw=black!55, fill=white, minimum width=2.52cm, minimum height=2.52cm, inner sep=0.5pt, align=center] (R4G) at ({1.5*\dx},\yrowfour) {\includegraphics[width=2.44cm,height=2.44cm,keepaspectratio]{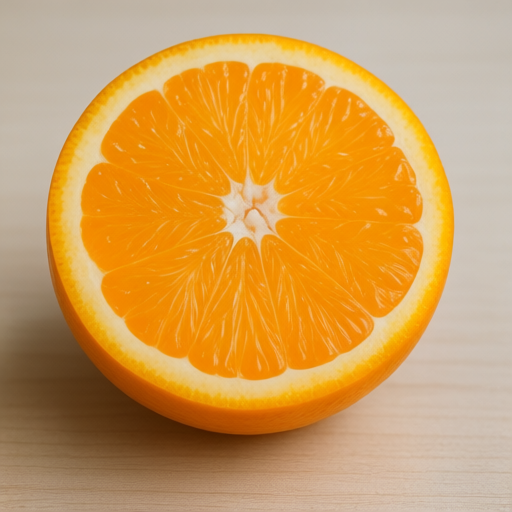}};
\node[draw=black!55, fill=white, minimum width=2.52cm, minimum height=2.52cm, inner sep=0.5pt, align=center] (R4C2) at ({2.5*\dx},\yrowfour) {\includegraphics[width=2.44cm,height=2.44cm,keepaspectratio]{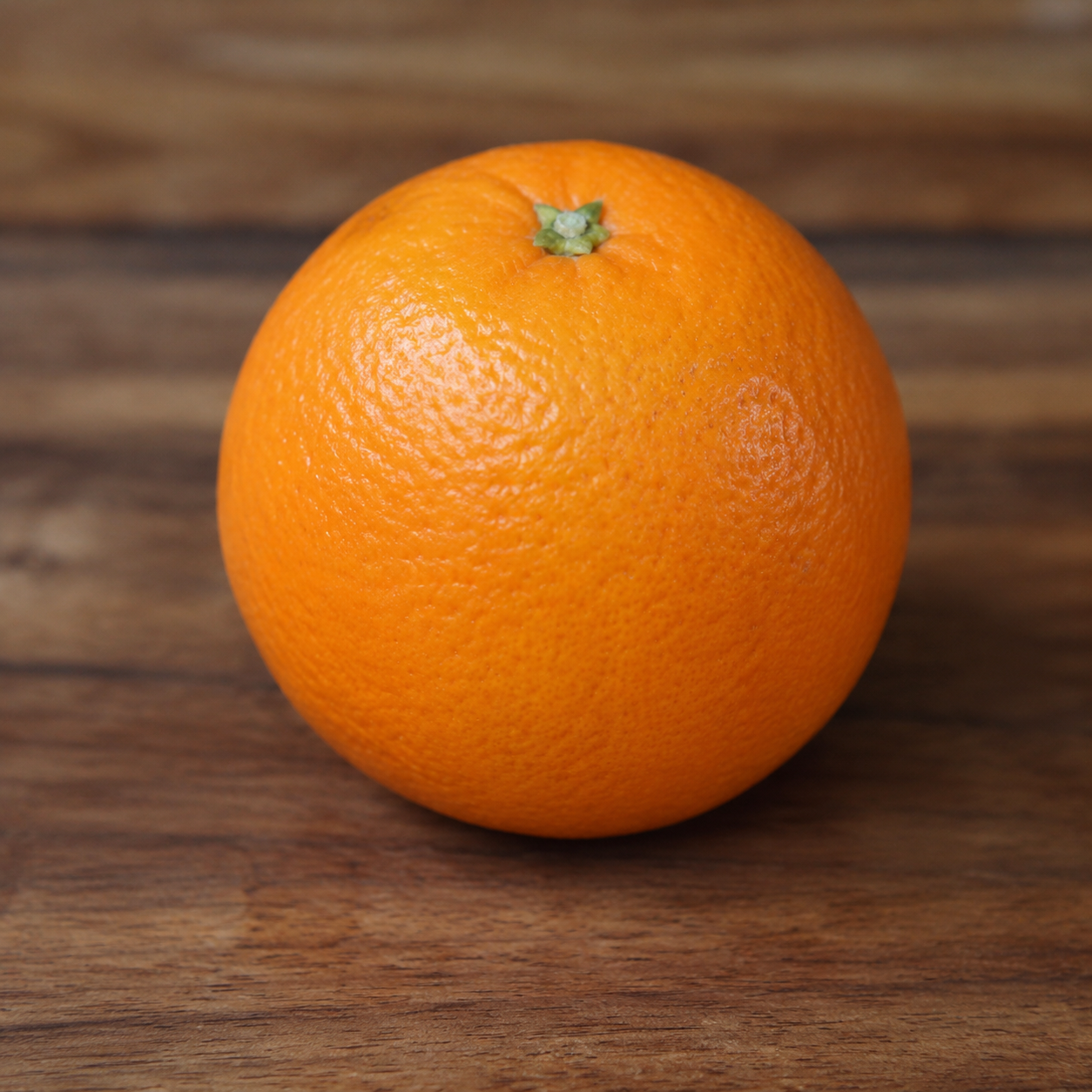}};
\node[draw=black!55, fill=white, minimum width=2.52cm, minimum height=2.52cm, inner sep=1pt, text width=2.30cm, align=left, font=\tiny] (R4C3) at ({3.5*\dx},\yrowfour) {A high-fidelity photo of a white orange, centered, natural light, clean composition\,\ldots\ smooth peel with subtle highlights, plain neutral backdrop for clarity and depth.};
\node[draw=black!55, fill=white, minimum width=2.52cm, minimum height=2.52cm, inner sep=0.5pt, align=center] (R4C4) at ({4.5*\dx},\yrowfour) {\includegraphics[width=2.44cm,height=2.44cm,keepaspectratio]{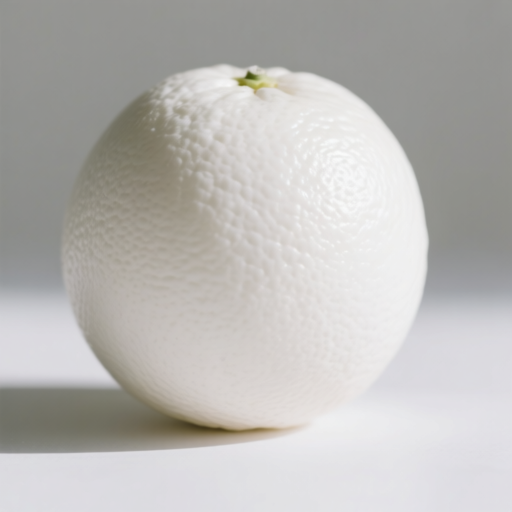}};
\node[anchor=north west, align=left, font=\scriptsize\rmfamily, inner sep=0pt] at ($(R4C1.south west)+(0,-0.04)$) {\mbox{Prompt: ``a photo of a white orange''}};
\node[draw, dashed, line width=0.9pt, rounded corners=2.5pt, inner xsep=1.5pt, inner ysep=2.5pt, xshift=0.10cm, fit=(H3)(H4)(R1C3)(R1C4)(R2C3)(R2C4)(R3C3)(R3C4)(R4C3)(R4C4)] {};
\end{tikzpicture}%
\end{adjustbox}
\caption{\textbf{Qualitative comparison on four prompts.} From left to right: images from the original prompt (Qwen-Image), GRPO, and gpt-image~2, followed by the rewrite snippet and the \textbf{SCOPE} output (Qwen-Image).}
\label{fig:qual_compositional}
\end{figure}

Figure~\ref{fig:qual_compositional} targets prompts that contradict common visual priors.
With the original prompt, the frozen Qwen-Image places the cup beneath the couch, draws five
handbags instead of four, and renders an ordinary orange. GRPO still misses the layout, the
count (three handbags), and the color. The rewriter trained with SCOPE spells out each
constraint explicitly: the vertical layout (\textit{``couch lower, cup upper''}), one object
per position (\textit{``four handbags in a neat row\ldots{} each on a white pedestal''}), and
the counterfactual color. Because SCOPE credits complete semantic spans during training, the
rewriter learns to keep each constraint as an intact phrase rather than scattering it across
subword tokens. This lifts a frozen open-weight Qwen-Image to the level of the commercial
\texttt{gpt-image 2} on layout and counting. On counterfactual attributes, SCOPE goes further: it is only model that renders a white orange, while \texttt{gpt-image 2} falls back to an orange-colored prior.

%% file: content/sec/5_conclusion.tex
\section{Conclusion}
\label{sec:conclusion}

We present \textbf{SCOPE}, a semantic-aware credit optimization framework for RL-based prompt engineering with frozen text-to-image generators. By pooling policy log-probability shifts within coherent semantic spans to modulate a shared completion advantage, SCOPE resolves subword credit fragmentation while requiring no separate span rewards, additional image renders, or external parsers. Across Qwen-Image, SD3, and FLUX.1-dev, SCOPE achieves substantial improvements over vanilla GRPO, delivering average relative gains of $9.48\%$ on GenEval and $7.62\%$ on T2I-CompBench. Accordingly, SCOPE establishes structure-aligned credit assignment as a plug-and-play paradigm for unlocking the compositional potential of frozen visual generators.

%% file: content/sec/append.tex
\raggedbottom
\setcounter{section}{1}

\subsection{Implementation overview}
\label{sec:app_impl_overview}

Distributed policy training optimizes the Qwen3-8B rewriter, while separate reward and rendering workers serve the frozen T2I backbones.
Qwen3-8B generates the spans during the rollout, without a separate span model or parser service.
Weighting by policy shifts within spans follows Eqs.~\ref{eq:span_log_shift} to~\ref{eq:span_normalized_weight} in the main text.
The worker layout and hardware placement do not change the objective or evaluation protocol.

\subsection{Hyperparameters}
\label{sec:app_hyper}

\begin{table}[H]
\centering
\small
\caption{\textbf{Optimization configuration for prompt rewriting, span weighting, and normalization.}}
\label{tab:hyper}
\begin{tabular}{ll}
\toprule
\textbf{Component} & \textbf{Value} \\
\midrule
Policy model & Qwen3-8B \\
Image models & Qwen-Image, SD3, FLUX.1-dev (frozen, Sec.~\ref{sec:experiments}) \\
Span acquisition & Qwen3-8B jointly generates nonempty spans separated by \textnormal{\textbar} \\
Training data & 10k selected GenEval + 4.2k T2I-CB prompts \\
Grouped samples per prompt $N$ & 8 \\
Learning rate & $1 \times 10^{-6}$ \\
KL coefficient $\beta_{\mathrm{KL}}$ & 0.01 \\
Max completion length & 512 \\
Span shift $b_{g,i,s}$ & mean log probability shift over valid span tokens (current minus old) \\
Span raw weight $v_{g,i,s}$ & $\operatorname{clip}(\exp(\beta b_{g,i,s}),0.8,1.2)$ \\
Span normalization & mean weighted by valid token counts $=1$, detached \\
\bottomrule
\end{tabular}
\end{table}

\paragraph{Launcher defaults vs.\ paper.}
The reported runs use the configuration in Table~\ref{tab:hyper}. Local launcher defaults can differ.
To reproduce the tables exactly, use this configuration and the same data and sampling protocol.

\subsection{Rendering configuration}
\label{sec:app_t2i_gen}

Table~\ref{tab:t2i_gen} specifies the rendering configuration for training rewards and benchmark evaluation.

\begin{table}[H]
\centering
\small
\caption{\textbf{Fixed rendering configuration for training rewards and benchmark evaluation in our runs.}}
\label{tab:t2i_gen}
\begin{tabular}{llll}
\toprule
\textbf{Backbone} & \textbf{Resolution} & \textbf{Steps} & \textbf{CFG scale} \\
\midrule
Qwen-Image & $512{\times}512$ & 20 & 4.0 \\
FLUX.1-dev & $512{\times}512$ & 20 & 3.5 \\
Stable Diffusion~3 (SD3) & $512{\times}512$ & 28 & 7.5 \\
\bottomrule
\end{tabular}
\end{table}

\subsection{Span Extraction and Fallbacks}
\label{subsec:span_implementation}
\label{sec:app_fallbacks}

The rewrite instruction preserves the request and clarifies subjects, counts, colors, attributes, and spatial relations.
The rewrite and its boundaries are generated in one completion, with \textnormal{\textbar} appearing only between adjacent nonempty spans.
The separator is not a generation stop token.
Tokenizer character offsets align each extracted span with its rollout tokens, keeping the partition consistent with the policy loss.

A valid partition requires a valid answer wrapper, nonempty spans, and full coverage of valid answer content tokens with consistent offsets and token counts.
SCOPE uses uniform GRPO token weights for malformed wrappers, empty or missing spans, and consecutive or trailing separators.
Truncation, incomplete coverage, alignment failures, and token count mismatches also trigger this fallback.
We handle one benign end of sequence (EOS) mismatch separately.
If the completion mask has exactly one extra trailing placeholder relative to the tokenized prompt text, we retain the aligned prefix.
The extra token receives zero weight. The aligned prefix determines the span statistics used to compute the shared coefficients.

Table~\ref{tab:app_fallback_rates} reports fallback counts and rates across benchmarks and frozen image backbones.

\begin{table}[H]
\centering
\small
\caption{\textbf{Fallback frequencies in the logged training rollouts by benchmark and frozen image backbone.}}
\label{tab:app_fallback_rates}
\begin{tabular}{llrrr}
\toprule
\textbf{Benchmark} & \textbf{Backbone} & \textbf{Fallbacks} & \textbf{Rollouts} & \textbf{Rate} \\
\midrule
GenEval & FLUX.1-dev & 147 & 80,000 & 0.18\% \\
 & SD3 & 165 & 80,000 & 0.21\% \\
 & Qwen-Image & 89 & 80,000 & 0.11\% \\
\midrule
T2I-CompBench & FLUX.1-dev & 269 & 33,600 & 0.80\% \\
 & SD3 & 247 & 33,600 & 0.74\% \\
 & Qwen-Image & 249 & 33,600 & 0.74\% \\
\bottomrule
\end{tabular}
\end{table}

\subsection{Z-Image Rewriter}
\label{sec:app_training_free_scores}

These tables report the Z-Image rewriter scores for each frozen backbone under the evaluation protocol in Sec.~\ref{sec:experiments}.
They reproduce the corresponding rows of the main tables with the same rendering configuration.

\begin{table}[H]
\centering
\footnotesize
\setlength{\tabcolsep}{3.2pt}
\renewcommand{\arraystretch}{1.0}
\caption{\textbf{GenEval scores for the Z-Image rewriter across the three frozen backbones.} Subset of Table~\ref{tab:geneval}.}
\label{tab:app_geneval_training_free}
\begin{tabular}{l c c c c c c c}
\toprule
\textbf{Backbone} & \textbf{1 Obj.} & \textbf{2 Obj.} & \textbf{Cnt.} & \textbf{Clr.} & \textbf{Pos.} & \textbf{Attr.} & \textbf{Avg.}$\uparrow$ \\
\midrule
Qwen-Image & 0.95 & 0.89 & 0.69 & 0.82 & 0.51 & 0.73 & 0.76 \\
SD3 & 0.58 & 0.54 & 0.37 & 0.42 & 0.25 & 0.37 & 0.42 \\
FLUX.1-dev & 0.53 & 0.50 & 0.56 & 0.42 & 0.30 & 0.36 & 0.44 \\
\bottomrule
\end{tabular}
\end{table}

\begin{table}[H]
\centering
\small
\setlength{\tabcolsep}{3.5pt}
\renewcommand{\arraystretch}{1.05}
\caption{\textbf{T2I-CompBench scores for the Z-Image rewriter by backbone.} Subset of Table~\ref{tab:benchmark_summary_across_backbones}.}
\label{tab:app_t2i_training_free}
\begin{tabular}{l c c c c c c}
\toprule
\textbf{Backbone} & \textbf{Color} & \textbf{Shape} & \textbf{Texture} & \textbf{Spatial} & \textbf{Numeracy} & \textbf{Complex} \\
\midrule
Qwen-Image & 0.8324 & 0.5930 & 0.7285 & 0.3880 & 0.6963 & 0.3922 \\
SD3 & 0.7933 & 0.5846 & 0.7141 & 0.2977 & 0.6286 & 0.3876 \\
FLUX.1-dev & 0.7624 & 0.5363 & 0.6396 & 0.3162 & 0.6492 & 0.3817 \\
\bottomrule
\end{tabular}
\end{table}

\subsection{Compute}
\label{sec:app_compute}

Policy optimization and auxiliary jobs ran on NVIDIA H200 GPUs.
Exact node counts depend on the notes for each deployment.
Table~\ref{tab:t2i_gen} specifies sampling for each backbone without prescribing the cluster configuration.

\subsection{Environment variables and span weighting}
\label{sec:app_env}

Reward services expose endpoints, timeouts, generation count (default 8), and reward and advantage weights.
Span weighting reuses the current and old policy log probabilities from the clipped surrogate.
Its sensitivity $\beta$ is distinct from the KL coefficient $\beta_{\mathrm{KL}}$. 

\subsection{Rewrite instruction}

The system message asks for exactly \textnormal{\textless answer\textgreater span 1 \textbar{} span 2 \textbar{} span 3\textless/answer\textgreater}.
Before rendering and computation of image rewards, the trainer extracts the answer field, removes every delimiter, and concatenates the span text into one natural language prompt.
The format reward still evaluates the structured output.

\subsection{Reward notes}

Advantages follow the hybrid rule in Eq.~\ref{eq:hybrid_advantage}.
During training, each group contains $N=8$ prompt members that share one random diffusion seed.
This controls renderer noise when comparing rewrites within each group.
Groups draw seeds independently, and training and evaluation use fresh seeds across prompts and outer loops.
Section~\ref{subsec:exp_setup} details the data splits and evaluation protocol.
Seed sharing is restricted to training groups.
The hybrid rule and shared seeds require no separate reward model or estimation module.

\subsection{RePrompt: Qwen-Image reproduction}
\label{sec:app_t2i_projection}
\citet{wu2026reprompt} do not report Qwen-Image results in their tables.
The RePrompt$^\dag$ entries for Qwen-Image in Tables~\ref{tab:geneval} and~\ref{tab:benchmark_summary_across_backbones} come from \emph{our reproduction} of their training and evaluation recipe.
These runs use the same reward clients, rendering seeds, and benchmark harness as our GRPO and \method{} runs.
For SD3 and FLUX.1-dev, the RePrompt rows reproduce the scores reported by \citet{wu2026reprompt}.

\subsection{Training procedure}
\label{sec:app_training_procedure}

\begin{algorithm}[H]
\caption{\textbf{SCOPE training procedure.}
The renderer $f_\phi$ and the evaluator stay frozen.
This step optimizes Eq.~\ref{eq:rl_objective} through Eq.~\ref{eq:outer_loss}.
Defaults are in Table~\ref{tab:hyper}. Span fallbacks are in Sec.~\ref{sec:app_fallbacks}.}
\label{alg:scope}
\newcommand{\eqrow}[2]{%
  \STATE\leavevmode
  \hbox to \linewidth{\ignorespaces #1\hfill\makebox[6.6em][l]{\textbf{\# Eq.~(\ref{#2})}}}%
}
\begin{algorithmic}[1]
\makeatletter
\renewenvironment{ALC@g}{%
  \begin{list}{\ALC@lno}{\itemsep 0.22em \itemindent\z@
    \listparindent\z@ \rightmargin\z@
    \topsep 0.08em \partopsep\z@ \parskip\z@ \parsep\z@
    \leftmargin\algorithmicindent
    \addtolength{\ALC@tlm}{\leftmargin}}%
}{\end{list}}
\makeatother
\setlength{\itemsep}{0.22em}
\setlength{\parsep}{0pt}
\setlength{\topsep}{0.1em}
\REQUIRE minibatch $\mathcal{B}$ from $\mathcal{D}$; rewriter $\pi_\theta$; frozen renderer $f_\phi$
\STATE On the first step set $\pi_{\mathrm{old}}\leftarrow\pi_\theta$. Afterward $\pi_{\mathrm{old}}$ is the pre-update snapshot.
\FOR{training step $s=1,2,\ldots$}
  \FOR{each prompt $x_g\in\mathcal{B}$}
    \STATE Sample $y_{g,1},\ldots,y_{g,N-1}\sim\pi_\theta(\cdot\mid x_g)$
    \STATE Set $y_{g,N}\leftarrow x_g$, so $c(y_{g,N})=x_g$, and $A_{g,N}\leftarrow 0$
    \STATE Share one rendering seed across the $N$ group members
    \FOR{$i=1,\ldots,N$}
      \eqrow{$r_{g,i}\leftarrow w_{\mathrm{t2i}}\,r_{\mathrm{t2i}}+w_{\mathrm{fmt}}\,r_{\mathrm{fmt}}$}{eq:reward_mix}
    \ENDFOR
    \eqrow{$A_{g,i}$ from $b_g=r_{g,N}$, with $\mu_g$ and $\hat\sigma_g$ over $i<N$}{eq:hybrid_advantage}
    \FOR{$i=1,\ldots,N-1$}
      \eqrow{$\Delta_{g,i,t}\leftarrow\log\pi_\theta-\log\pi_{\mathrm{old}}$}{eq:span_log_shift}
      \eqrow{$\rho_{g,i,t}\leftarrow\exp(\Delta_{g,i,t})$}{eq:rho_def}
      \IF{the answer partition aligns to the rollout tokens}
        \eqrow{$b_{g,i,s}\leftarrow\mathrm{mean}_{t\in\mathcal{T}_{g,i,s}}\Delta_{g,i,t}$}{eq:span_local_weight}
        \eqrow{$v_{g,i,s}\leftarrow\mathrm{clip}(\exp(\beta b_{g,i,s}),0.8,1.2)$}{eq:span_coefficient}
        \eqrow{$w_{g,i,s}\leftarrow v_{g,i,s}/\mathrm{mean}(v)$}{eq:span_normalized_weight}
        \eqrow{$\widetilde{A}_{g,i,t}\leftarrow A_{g,i}\,\mathrm{stopgrad}(w_{g,i,s})$}{eq:span_modulated_advantage}
        \eqrow{\mbox{$\ell^{\mathrm{scope}}_{g,i}$}}{eq:scope_surrogate}
      \ELSE
        \eqrow{\mbox{$\ell^{\mathrm{uni}}_{g,i}$}}{eq:grpo_surrogate}
      \ENDIF
    \ENDFOR
  \ENDFOR
  \eqrow{$\mathcal{L}_{\mathrm{pol}}\leftarrow |\mathcal{B}|^{-1}\sum_g (N-1)^{-1}\sum_{i<N}\ell_{g,i}$}{eq:outer_loss}
  \STATE Add the optional KL penalty against the frozen reference, which is distinct from $\pi_{\mathrm{old}}$, and update $\theta$
  \STATE Set $\pi_{\mathrm{old}}\leftarrow\pi_\theta$ for the next step
\ENDFOR
\RETURN $\pi_\theta$
\end{algorithmic}
\end{algorithm}

\input{content/sec/x_appendix_qualitative}

%% file: content/sec/x_appendix_qualitative.tex
\subsection{Additional Qualitative Comparisons}
\label{sec:app_extra_qual}

\begin{figure}[H]
\centering
\footnotesize
\sffamily
\setlength{\fboxsep}{0pt}%
\begin{adjustbox}{max width=\linewidth, max height=0.70\textheight, center}
\begin{tikzpicture}[x=1cm,y=1cm]
\tikzset{%
  hdr/.style={minimum width=2.62cm, align=center, font=\bfseries\footnotesize},
}
\def\dx{2.72}
\pgfmathsetmacro{\rowgap}{3.48}
\pgfmathsetmacro{\yone}{-2.08}
\node[hdr] (H1) at ({0.5*\dx},0.28) {User prompt};
\node[hdr] (HG) at ({1.5*\dx},0.28) {GRPO};
\node[hdr, text width=2.62cm] (H2) at ({2.5*\dx},0.28) {gpt-image~2};
\node[hdr] (H3) at ({3.5*\dx},0.28) {Rewrite / rationale};
\node[hdr] (H4) at ({4.5*\dx},0.28) {\method{} (Ours)};
\pgfmathsetmacro{\y}{\yone}
\node[draw=black!55, fill=white, minimum width=2.72cm, minimum height=2.72cm, inner sep=0.5pt, align=center] (R1C1) at ({0.5*\dx},\y) {\includegraphics[width=2.66cm,height=2.66cm,keepaspectratio]{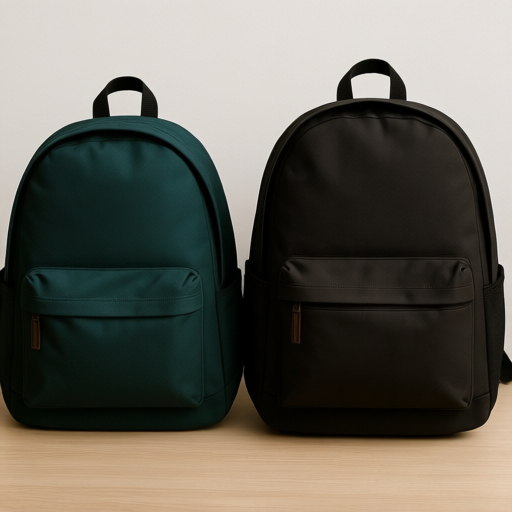}};
\node[draw=black!55, fill=white, minimum width=2.72cm, minimum height=2.72cm, inner sep=0.5pt, align=center] (R1G) at ({1.5*\dx},\y) {\includegraphics[width=2.66cm,height=2.66cm,keepaspectratio]{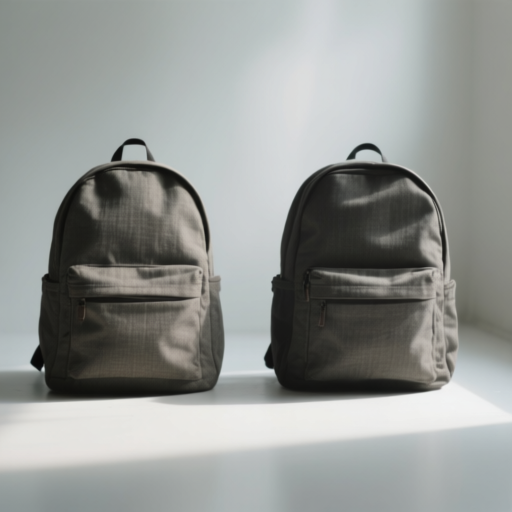}};
\node[draw=black!55, fill=white, minimum width=2.72cm, minimum height=2.72cm, inner sep=0.5pt, align=center] (R1C2) at ({2.5*\dx},\y) {\includegraphics[width=2.66cm,height=2.66cm,keepaspectratio]{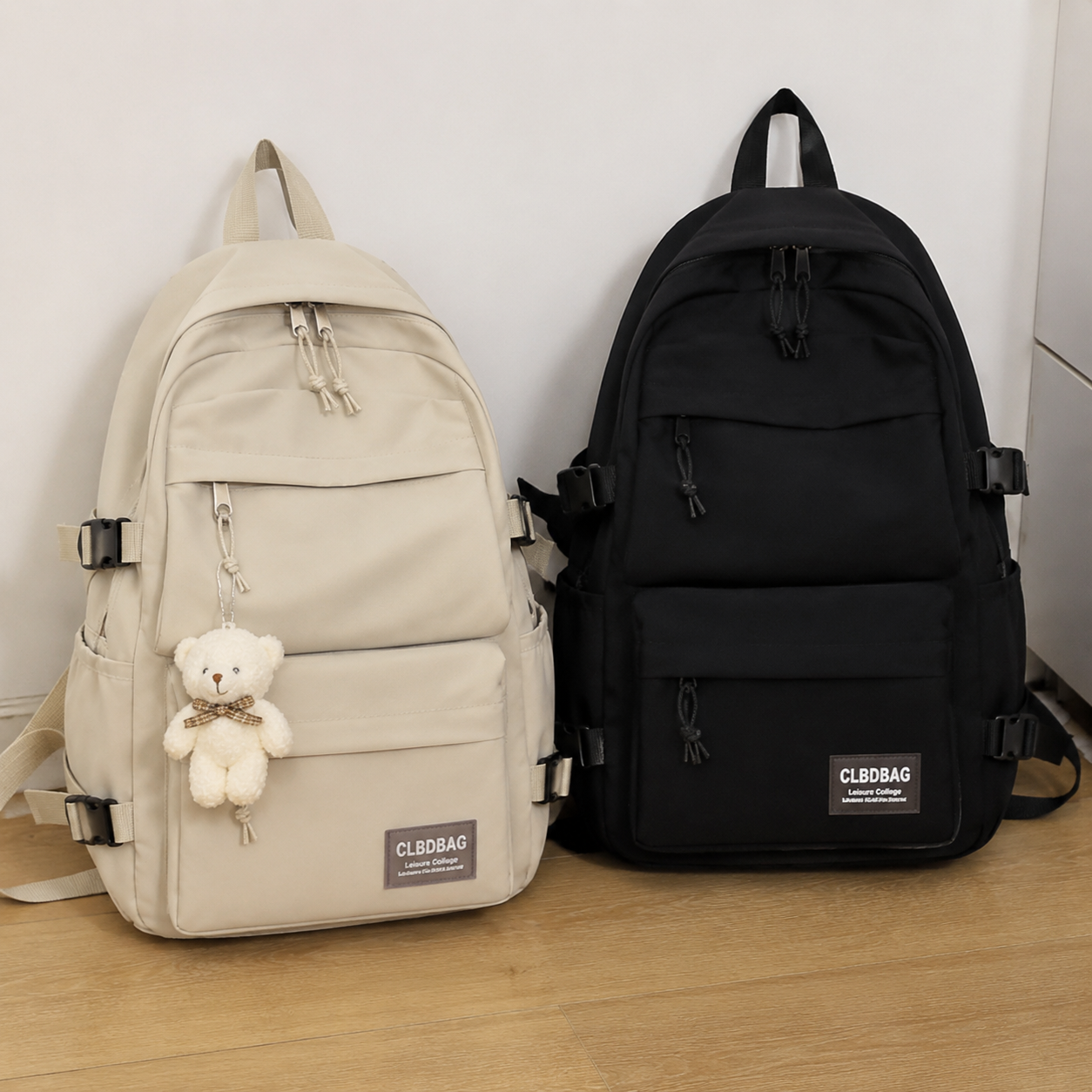}};
\node[draw=black!55, fill=white, minimum width=2.72cm, minimum height=2.72cm, inner sep=2pt, text width=2.52cm, align=left, font=\scriptsize] (R1C3) at ({3.5*\dx},\y) {A photo of two backpacks, each on the left and right sides of the frame, with detailed textures and muted earth tones\,\ldots\ soft natural lighting, shallow DoF, emphasizing the two objects.};
\node[draw=black!55, fill=white, minimum width=2.72cm, minimum height=2.72cm, inner sep=0.5pt, align=center] (R1C4) at ({4.5*\dx},\y) {\includegraphics[width=2.66cm,height=2.66cm,keepaspectratio]{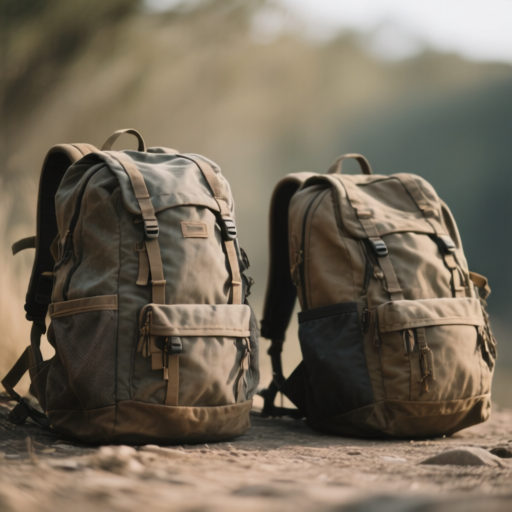}};
\node[anchor=north west, align=left, font=\footnotesize\rmfamily, inner sep=0pt] at ($(R1C1.south west)+(0,-0.08)$) {\mbox{Prompt: ``a photo of two backpacks''}};
\pgfmathsetmacro{\yrowtwo}{\yone-\rowgap}
\node[draw=black!55, fill=white, minimum width=2.72cm, minimum height=2.72cm, inner sep=0.5pt, align=center] (R2C1) at ({0.5*\dx},\yrowtwo) {\includegraphics[width=2.66cm,height=2.66cm,keepaspectratio]{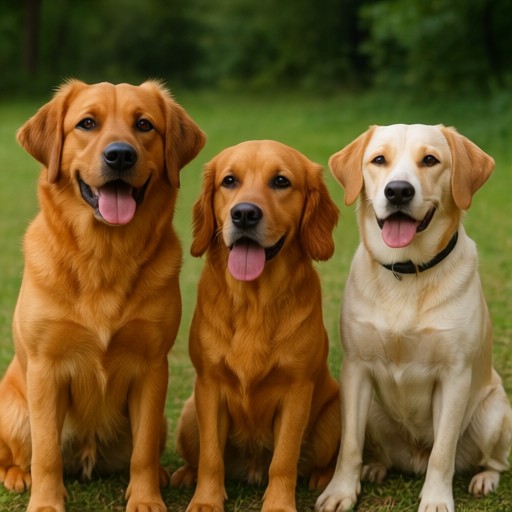}};
\node[draw=black!55, fill=white, minimum width=2.72cm, minimum height=2.72cm, inner sep=0.5pt, align=center] (R2G) at ({1.5*\dx},\yrowtwo) {\includegraphics[width=2.66cm,height=2.66cm,keepaspectratio]{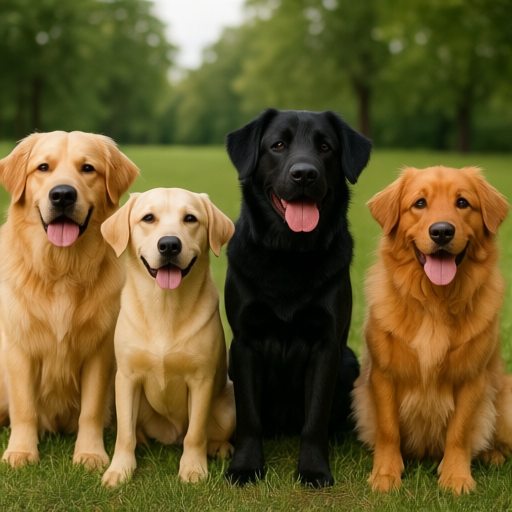}};
\node[draw=black!55, fill=white, minimum width=2.72cm, minimum height=2.72cm, inner sep=0.5pt, align=center] (R2C2) at ({2.5*\dx},\yrowtwo) {\includegraphics[width=2.66cm,height=2.66cm,keepaspectratio]{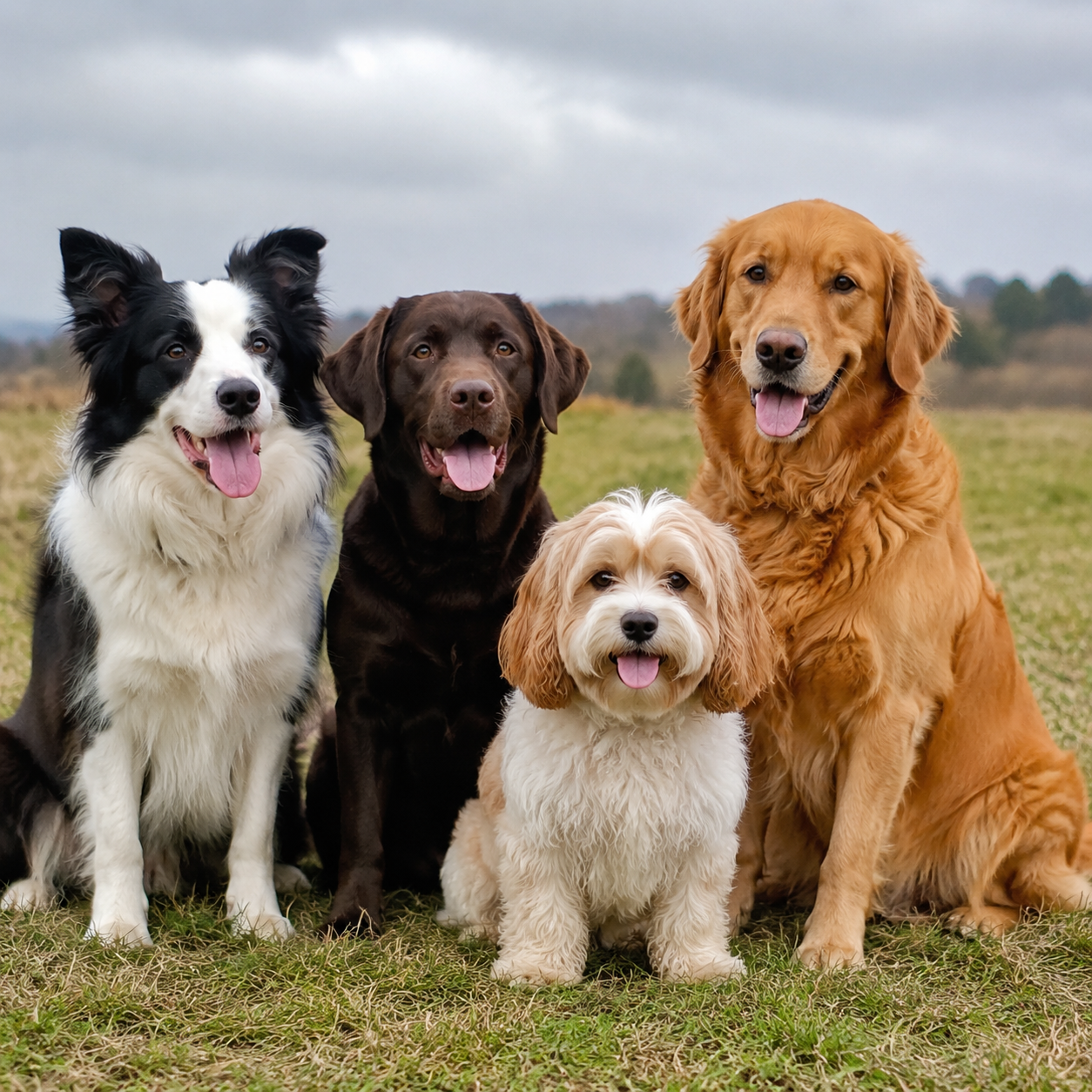}};
\node[draw=black!55, fill=white, minimum width=2.72cm, minimum height=2.72cm, inner sep=2pt, text width=2.52cm, align=left, font=\scriptsize] (R2C3) at ({3.5*\dx},\yrowtwo) {A photo of four dogs, spaced left, center-left, center-right, and right in the frame, outdoor setting\,\ldots\ soft light on fur and faces, balanced composition.};
\node[draw=black!55, fill=white, minimum width=2.72cm, minimum height=2.72cm, inner sep=0.5pt, align=center] (R2C4) at ({4.5*\dx},\yrowtwo) {\includegraphics[width=2.66cm,height=2.66cm,keepaspectratio]{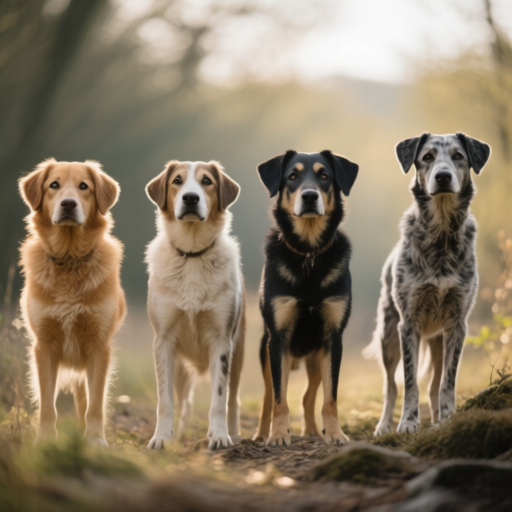}};
\node[anchor=north west, align=left, font=\footnotesize\rmfamily, inner sep=0pt] at ($(R2C1.south west)+(0,-0.08)$) {\mbox{Prompt: ``a photo of four dogs''}};
\pgfmathsetmacro{\yrowthree}{\yrowtwo-\rowgap}
\node[draw=black!55, fill=white, minimum width=2.72cm, minimum height=2.72cm, inner sep=0.5pt, align=center] (R3C1) at ({0.5*\dx},\yrowthree) {\includegraphics[width=2.66cm,height=2.66cm,keepaspectratio]{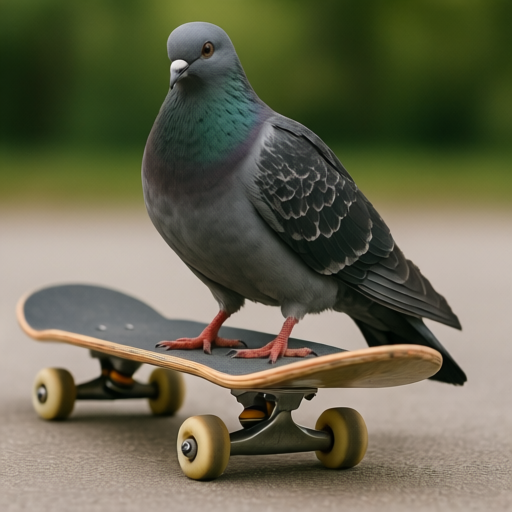}};
\node[draw=black!55, fill=white, minimum width=2.72cm, minimum height=2.72cm, inner sep=0.5pt, align=center] (R3G) at ({1.5*\dx},\yrowthree) {\includegraphics[width=2.66cm,height=2.66cm,keepaspectratio]{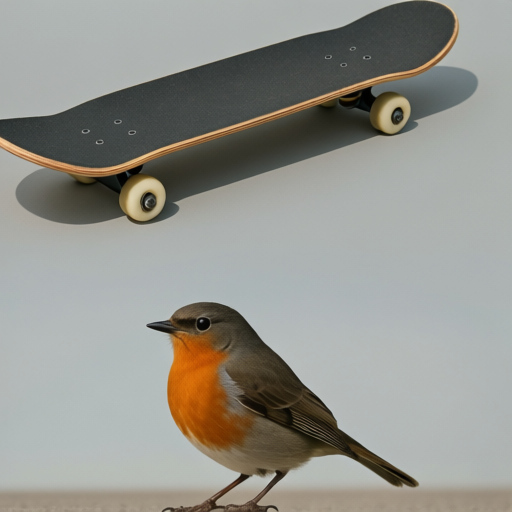}};
\node[draw=black!55, fill=white, minimum width=2.72cm, minimum height=2.72cm, inner sep=0.5pt, align=center] (R3C2) at ({2.5*\dx},\yrowthree) {\includegraphics[width=2.66cm,height=2.66cm,keepaspectratio]{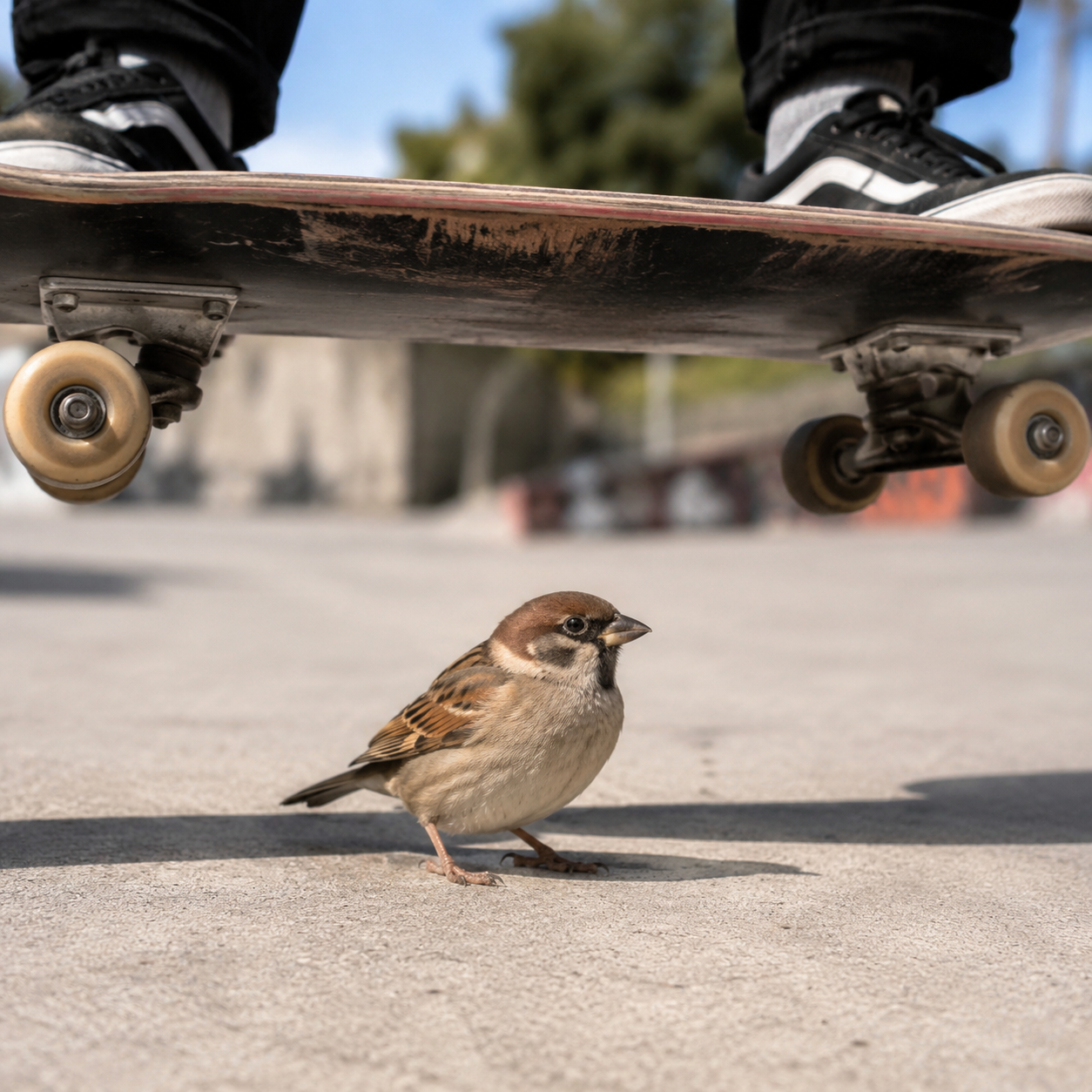}};
\node[draw=black!55, fill=white, minimum width=2.72cm, minimum height=2.72cm, inner sep=2pt, text width=2.52cm, align=left, font=\scriptsize] (R3C3) at ({3.5*\dx},\yrowthree) {A photo of a bird below a skateboard on the ground, board low in the frame, bird in the lower half looking up\,\ldots\ spatial layout emphasized.};
\node[draw=black!55, fill=white, minimum width=2.72cm, minimum height=2.72cm, inner sep=0.5pt, align=center] (R3C4) at ({4.5*\dx},\yrowthree) {\includegraphics[width=2.66cm,height=2.66cm,keepaspectratio]{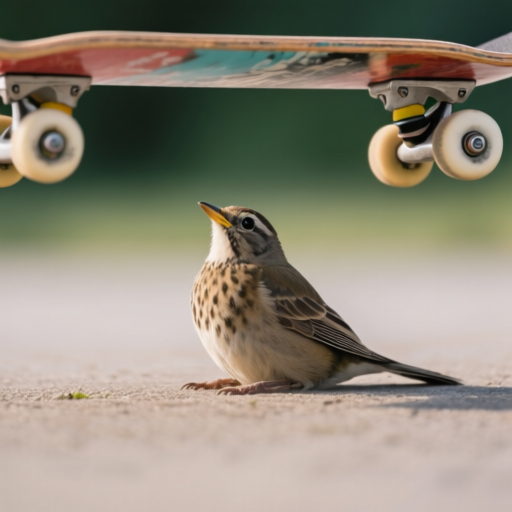}};
\node[anchor=north west, align=left, font=\footnotesize\rmfamily, inner sep=0pt] at ($(R3C1.south west)+(0,-0.08)$) {\mbox{Prompt: ``a photo of a bird below a skateboard''}};
\pgfmathsetmacro{\yrowfour}{\yrowthree-\rowgap}
\node[draw=black!55, fill=white, minimum width=2.72cm, minimum height=2.72cm, inner sep=0.5pt, align=center] (R4C1) at ({0.5*\dx},\yrowfour) {\includegraphics[width=2.66cm,height=2.66cm,keepaspectratio]{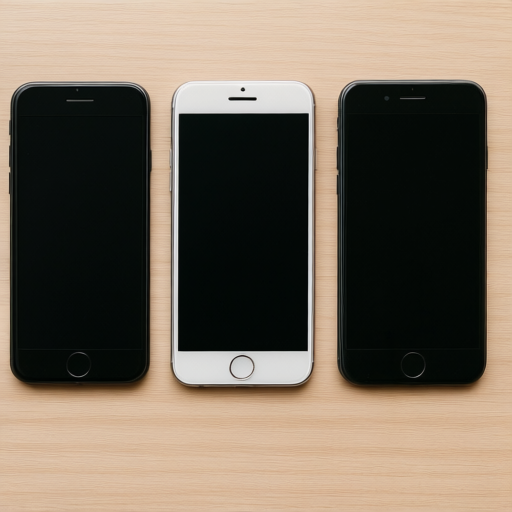}};
\node[draw=black!55, fill=white, minimum width=2.72cm, minimum height=2.72cm, inner sep=0.5pt, align=center] (R4G) at ({1.5*\dx},\yrowfour) {\includegraphics[width=2.66cm,height=2.66cm,keepaspectratio]{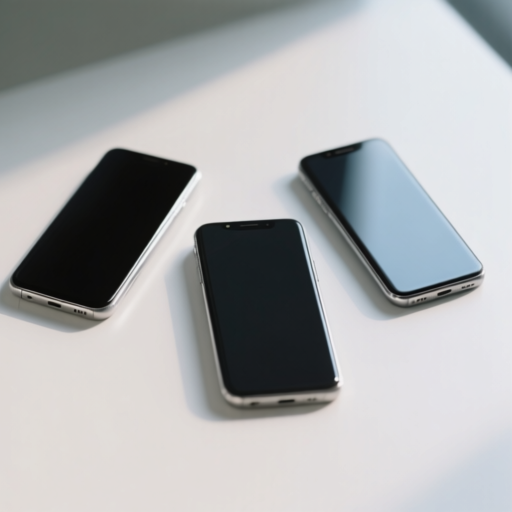}};
\node[draw=black!55, fill=white, minimum width=2.72cm, minimum height=2.72cm, inner sep=0.5pt, align=center] (R4C2) at ({2.5*\dx},\yrowfour) {\includegraphics[width=2.66cm,height=2.66cm,keepaspectratio]{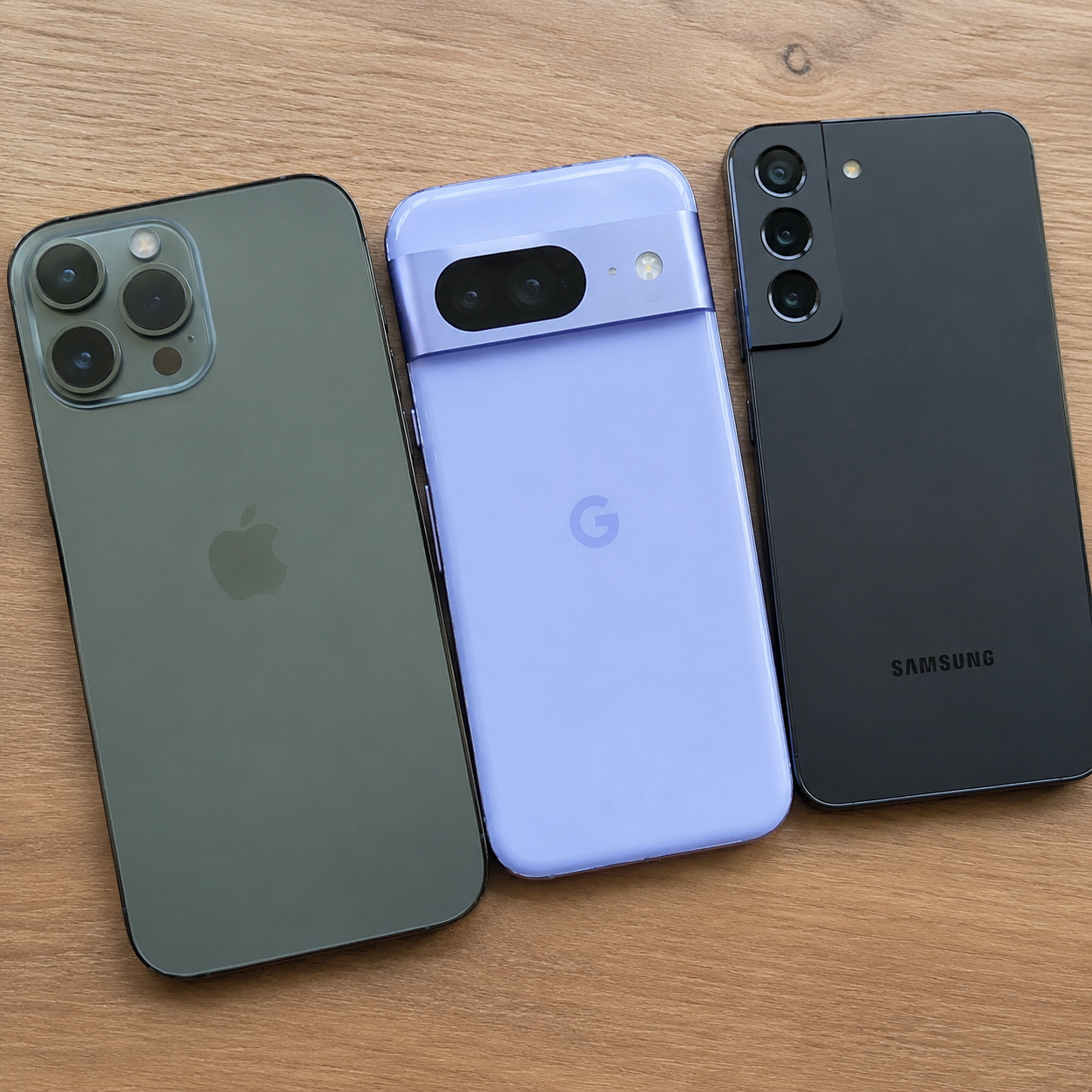}};
\node[draw=black!55, fill=white, minimum width=2.72cm, minimum height=2.72cm, inner sep=2pt, text width=2.52cm, align=left, font=\scriptsize] (R4C3) at ({3.5*\dx},\yrowfour) {Three cell phones in a loose triangle, left, center, and right\,\ldots\ minimalist backdrop, soft ambient light on sleek shells, calm modern mood.};
\node[draw=black!55, fill=white, minimum width=2.72cm, minimum height=2.72cm, inner sep=0.5pt, align=center] (R4C4) at ({4.5*\dx},\yrowfour) {\includegraphics[width=2.66cm,height=2.66cm,keepaspectratio]{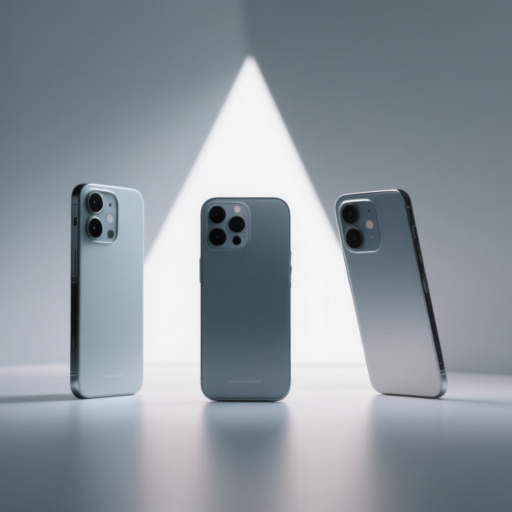}};
\node[anchor=north west, align=left, font=\footnotesize\rmfamily, inner sep=0pt] at ($(R4C1.south west)+(0,-0.08)$) {\mbox{Prompt: ``a photo of three cell phones''}};
\node[draw, dashed, line width=0.9pt, rounded corners=2.5pt, inner xsep=1.5pt, inner ysep=5pt, xshift=0.06cm, fit=(H3)(H4)(R1C3)(R1C4)(R2C3)(R2C4)(R3C3)(R3C4)(R4C3)(R4C4)] {};
\end{tikzpicture}%
\end{adjustbox}
\caption{\textbf{Additional qualitative comparisons on four prompts, using the same five columns as Figure~\ref{fig:qual_compositional}.}}
\label{fig:app_qual_extra_merged}
\end{figure}

Figure~\ref{fig:app_qual_extra_merged} uses the five columns in Figure~\ref{fig:qual_compositional}, with matched seeds within each row.
The dashed frame encloses rewrite snippets and SCOPE outputs. Rows are grouped by prompt type and evaluation setting.

